\documentclass{projectXtemplate}
\usepackage[utf8]{inputenc}
\usepackage[english]{babel}
\usepackage{graphicx}
\graphicspath{{./images/}}
\usepackage[bookmarks=false]{hyperref}
\usepackage[backend=biber, sorting=none]{biblatex}
\usepackage{upquote}
\usepackage{amsmath}
\usepackage{amssymb}
\usepackage{mathtools}
\usepackage{comment}
\usepackage{bm}
\pdfoutput=1

\bibliography{biblio}
  {\begin{list}{}%
          {\setlength{\leftmargin}{#1}}%
          \item[]%
  }
  {\end{list}}

\usepackage{calrsfs}%%%%%%%%%%<------add
\DeclareMathAlphabet{\pazocal}{OMS}{zplm}{m}{n}%%%%%%%%%%<------add

\title{Physics-Informed Machine Learning Simulator for Wildfire Propagation}

\author{\textbf{Luca Bottero*, Francesco Calisto*, Giovanni Graziano*, Valerio Pagliarino*,}}
\affil{\textbf{Martina Scauda\dag, Sara Tiengo* and Simone Azeglio*}}
\author{*Student, Department of Physics, University of Turin, \dag Student, Department of Mathematics, University of Turin}
\affil{E-mails: luca.bottero192, francesco.calisto, giovanni.graziano136, valerio.pagliarino,}
\affil{martina.scauda, sara.tiengo, simone.azeglio@edu.unito.it}

\thisvolume{XX}
\thismonth{October}
\thisyear{20XX}
\thisdoi{XX}
\thisstartpage{01}
\thisendpage{03}

\begin{document}

\maketitle
\thispagestyle{thefirstpage}

\begin{center}
   ABSTRACT 
\end{center} 

\vspace{-2mm}
\textbf{
The aim of this work is to evaluate the feasibility of re-implementing some key parts of the widely used Weather Research and Forecasting WRF-SFIRE simulator by replacing its core differential equations numerical solvers with state-of-the-art physics-informed machine learning techniques to solve ODEs and PDEs, in order to transform it into a real-time simulator for wildfire spread prediction. The main programming language used is Julia, a compiled language which offers better perfomance than interpreted ones, providing Just in Time (JIT) compilation with different optimization levels \cite{julia_perf}. Moreover, Julia is particularly well suited for numerical computation and for the solution of complex physical models, both considering the syntax and the presence of some specific libraries such as \textit{DifferentialEquations.jl} and \textit{ModellingToolkit.jl}.
}

\section{\textbf{INTRODUCTION}}
In recent years wildfires have been increasingly growing in intensity and frequency, becoming a serious threat to the health and socio-economic stability of various countries all over the world. In particular, Australia was devastated by the "Black Summer", the bushfire season between 2019 and 2020, and California just suffered from the most severe  wildfire season recorded in its modern history. According to the California Department of Forestry and Fire Protection \cite{cc_wf_CNN}: over 4 percent of its land was burned by more than 8,600 fires \cite{CALFIRE}.

The interconnection between wildfires and climate change is evident and has been studied by climate scientists over the years \cite{cc_wf}\cite{cc_wf_SCIENCE}.
On one hand, climate change favors the spread of wildfires. One of the main factors in play is the rise in global temperatures: the early beginning of spring leads to the rapid melting of snowpacks, causing land to dry out earlier and remain dry for longer. In addition, the hotter the air, the more water it soaks up from plants and soils; the "vapor pressure deficit" is used to measure the difference between how much water the air holds and how much it could hold: the higher this coefficient, the more soil and vegetation will dry out \cite{cc_wf_NATGEO}. Furthermore, bark beetles and other insects that survive at high temperatures are responsible for killing millions of trees, according to the Fourth National Climate Assessment \cite{FNCA}, turning them into kindling for wildfires. These two phenomena together produce an alarming growth in wildfire frequency and intensity or, following the terminology used in \cite{cc_wf}, the number of fires (NB) and the extension of the burnt area (BA).
On the other hand, wildfires impact climate change in two ways: firstly, they emit massive amounts of carbon dioxide and other pollutants that can affect regional and even global climate. Secondly, vast lands laid bare by fires are soaked with rain, increasing the risk for devastating landslides. On top of this, wildfires produce serious health risks due to the emission of toxic chemical compounds \cite{cc_wf_YALE}.

In such a worrisome scenario, having the support of fast and accurate wildfire spread simulators that can run \textbf{iteratively} in order to evaluate several different containment strategies is of unprecedented importance.

\subsection{\textbf{Real-time is revolutionary}}
The current stable version of WRF is able to produce highly accurate and validated predictions, but each simulation takes  several hours, which is acceptable for a single run. Nevertheless, if one wants to use the simulator for containment strategies, time is a major obstacle. In order to use WRF for that purpose, one should run the simulator iteratively with different initial scenarios, therefore predicting the outcome of various containment strategies and choosing the best one according to the output of the program. The revolutionary reach of such an approach is that it would limit human error in situations where taking the right decision quickly is extremely difficult. To reach this goal it is essential to significantly increase the computational efficiency of fire spread models, which can be done thanks to Machine Learning techniques.

Furthermore, in our proposal we will show that such techniques introduce a broad spectrum of other advantages, for example: the possibility to obtain a continuous solution for the problem instead of a discretized one on a grid, the possibility to forecast the evolution even outside the limits of the simulation domain, and the possibility to reconstruct the evolution back in time, with applications for forensic analysis.
The main programming language used in this work is \textit{Julia} for reasons explained below.

\subsection{\textbf{Physics-based or ML?}}
\textit{In the context of science, the well-known adage “a picture is worth a thousand words” might well be “a model is worth a thousand datasets.”}, writes Christopher Rackauckas in his trademark paper \cite{rackauckas}. A single sentence couldn't describe the true power of Scientific Machine Learning better. Machine Learning's utmost strength, i.e. universal flexibility to approximate any nonlinearity from data, as stated by the Universal Approximation Theorem \cite{UAT}, is at the same time its main drawback: the higher the complexity of the problem, the more data the algorithm needs to train.
While some fields can provide big data, such as bioinformatics \cite{bioinf}, many scientific disciplines are still limited in the use of Deep Learning (DL) by a lack of data. Another reason for the ineffectiveness of Machine Learning (ML) might be the apparent chaotic nature of a phenomenon. This is the case of wildfires and weather-related phenomena, which don't depend solely on the initial and boundary conditions, but evolve according to complex internal dynamics that cannot be predicted from mere data. In order to make a neural network learn such dynamics, one would need data with both high spatial and temporal resolution, something not available in this field \cite{tradeoff}.
For this reason, we think that the growing field of Scientific Machine Learning is a good compromise where climate and wildfire science could take advantage. The idea is to embed physical information, generally in the form of partial differential equations, into the ML model, thus constructing a Physics Informed Neural Network (PINN). A former physics-informed approach consists in training the model on data and constraining the solution space to the physically admissible ones (e.g. for incompressible fluid dynamics, any flow solution that breaks the mass conservation principle would not be considered by the model \cite{raissi}).
Another physics-informed architecture - the one considered in our case - consists instead in training the neural network directly on the physical model output data, employing the feed-forward networks as an efficient approximator of trajectories in the solution space. This approach leads to various advantages, such as avoiding the so-called \textit{curse of dimensionality}: as the dimensionality of an equation increases, the computational costs of a numerical equation solver grow exponentially, while PINNs can be proven to have a polynomial bound \cite{curse_dim}. This results in a significant speed up with respect to numerical solvers when solving high dimensional problems.
We decided to use the second technique since our goal is to evaluate the feasibility of a high efficiency, ML-based, wildfire spread simulator that uses the equations of the WRF-SFIRE model.

\vspace{1mm}
\section{\textbf{RESEARCH QUESTION}}
We aim to investigate the possibility to implement some key parts of the complex WRF-SFIRE physical model using an ML-based approach with all the advantages disclosed above and discussed in paragraph \ref{discussion}.\ref{advantages}.

\section{\textbf{METHODS}}

Firstly, we started by investigating the structure of the systems of partial differential equations (PDEs) that constitute the core of the physical model of WRF-SFIRE. In this way we detected the modules where replacing the current structure has the greatest impact and we found out how to re-implement them using our architecture.

\subsection{\textbf{WRF model exploration and profiling}}
\begin{figure}[ht!]
\begin {center}
\includegraphics[width=0.5\textwidth]{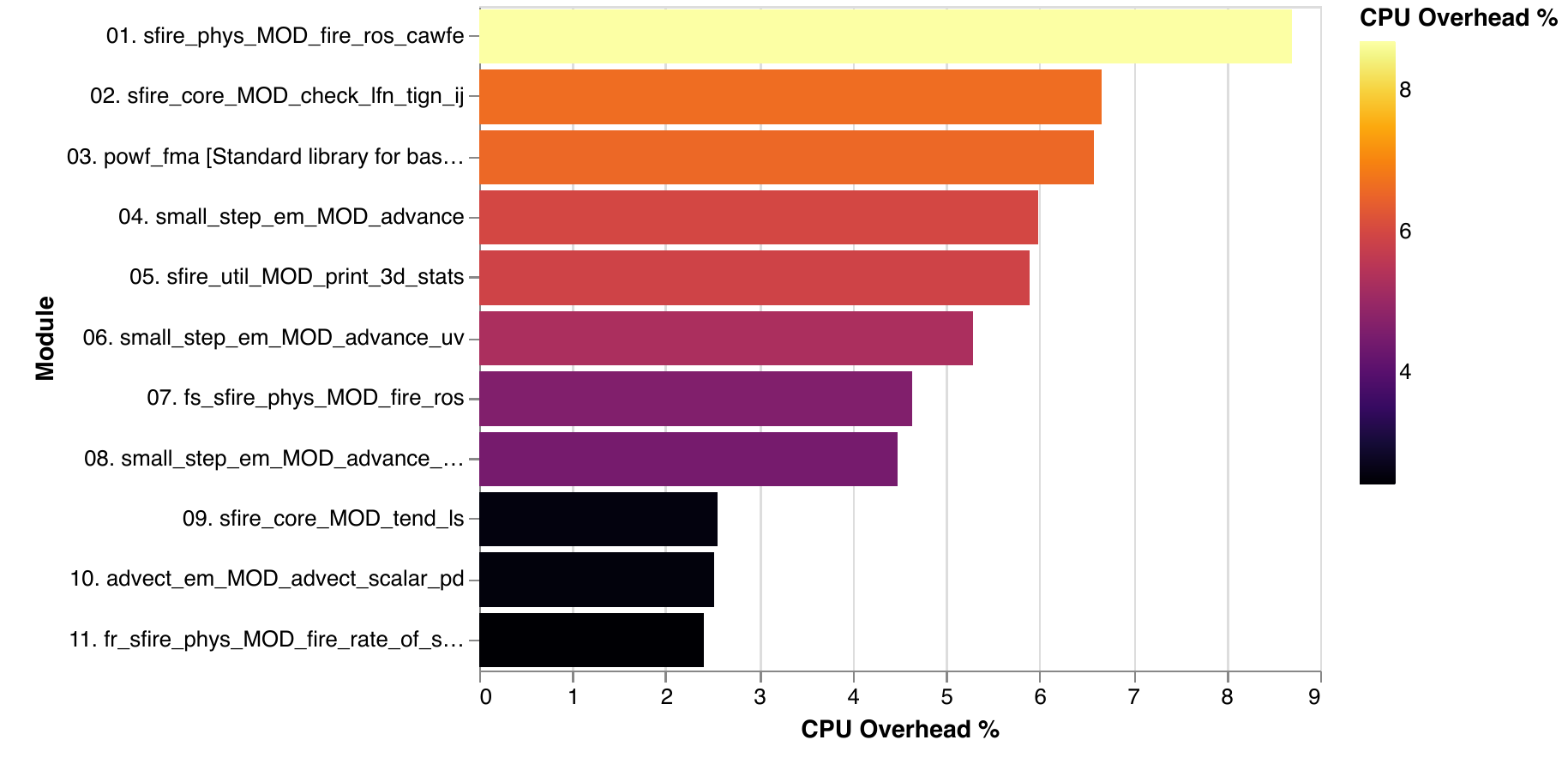}
\caption{Profiling of computation-expensive WRF modules}
\label{fig:profiling}
\end {center}
\end{figure}

In order to increase the computational efficiency of WRF-SFIRE model, we needed to identify the most computationally-intensive parts of WRF's code. We performed a profiling using the Linux \textit{perf} command \cite{PERFTOOL}. In this way, we obtained the CPU overhead of the subroutines measured running a variety of notable example scenarios provided by the developers of the WRF model. For comparison, we also ran a performance analysis for atmospheric simulations alone. The results we obtained are consistent with our hypothesis about what slows down the simulations: numerical solvers of the model's main differential equations are responsible for more than $20\%$ of the total overhead. The remaining machine time is due to I/O processes and other equations that require less complex computations or that are updated less frequently. The main difference between the purely atmospheric simulations and those with a fire simulation was, as expected, due to calculations of the fire spread rate and the resolution of the level set equation, which determines the fire propagation. This analysis was followed by an in-depth examination of the WRF's Fortran code, with particular attention to the SFIRE module, in order to understand how the problem of solving the governing equations is approached. We used this information to decide which equations to re-implement in our model.
This is our verdict: the level set equation, describing the evolution of fire area, is not only central from a conceptual point of view, but its numerical resolution has a significant impact on the CPU load during fire simulation. (Please see Fig.\ref{fig:profiling} and Fig.\ref{fig:profiling_list}). We were also interested in evaluating the possibility of applying PINNs to the resolution of the Euler system that is a system of 7 PDEs governing the atmosphere behavior. That is because even if the Euler system is not strictly linked to wildfire simulation, it has the greatest impact on the computational load and is physically coupled to the level set equation.

\textbf{Therefore, in this paper we will present a detailed application of the PINN architecture to the level set equation and a convergence study of PINNs training applied to the Euler system.}
\vspace{-6pt} 

\begin{table}[h]
\begin{center}
\begin{tabular}{ l  p{1.2cm} } \\
\hline \\
  \textbf{Module name (CPU, 4 threads)} &\hspace{-19pt}\textbf{Overhead} \% \\
01. sfire\_phys\_MOD\_fire\_ros\_cawfe  &         8.70 \\
02. sfire\_core\_MOD\_check\_lfn\_tign\_ij  &         6.66\\
03. powf\_fma\_[Standard library for basic math]  &     6.58\\
04. small\_step\_em\_MOD\_advance  &         5.99\\
05. sfire\_util\_MOD\_print\_3d\_stats  &         5.89\\
06. small\_step\_em\_MOD\_advance\_uv  &         5.29\\
07. fs\_sfire\_phys\_MOD\_fire\_ros  &         4.64\\
08. small\_step\_em\_MOD\_advance\_mu\_t  &         4.48\\
09. sfire\_core\_MOD\_tend\_ls  &         2.55\\
10. advect\_em\_MOD\_advect\_scalar\_pd  &         2.52\\
11. fr\_sfire\_phys\_MOD\_fire\_rate\_of\_spread  & 2.41 \\
\\\hline
\bottomrule
\end{tabular}
\caption{}
\label{fig:profiling_list}
\end{center}
\end{table}

\subsection{\textbf{A brief insight into the WRF-SFIRE model}}

Despite the focus of this project being on wildfire spread simulation, the current state-of-the-art fire spread models can produce realistic results only because they are integrated with a more complete atmospheric simulator; therefore presenting the structure of WRF-SFIRE is impossible without  insight into  the atmospheric model that is interrogated continuously by the fire propagation module.

\subsection{\textbf{Description of the physical model}}

The Weather Research and Forecasting (WRF) model is an atmospheric modeling system designed for Numerical Weather Prediction (NWP). The NWP procedure was designed by the meteorologist Vilherlm Bjerknes in 1904, originating from the assumption that weather forecasting is an initial value problem.
Therefore, once given the observed initial state of the atmosphere, the integration of governing equations is feasible and is assigned to the Advanced Research WRF (ARW), the dynamics solver of the model. However, the equations cannot be solved analytically, so they are described on a mesh and the computed solutions are an approximation of exact ones. A good and comprehensive description of the model is accessible in the Users' guide \cite{WRFGuide}.

\begin{figure}[H]
\begin {center}
\includegraphics[width=0.3\textwidth]{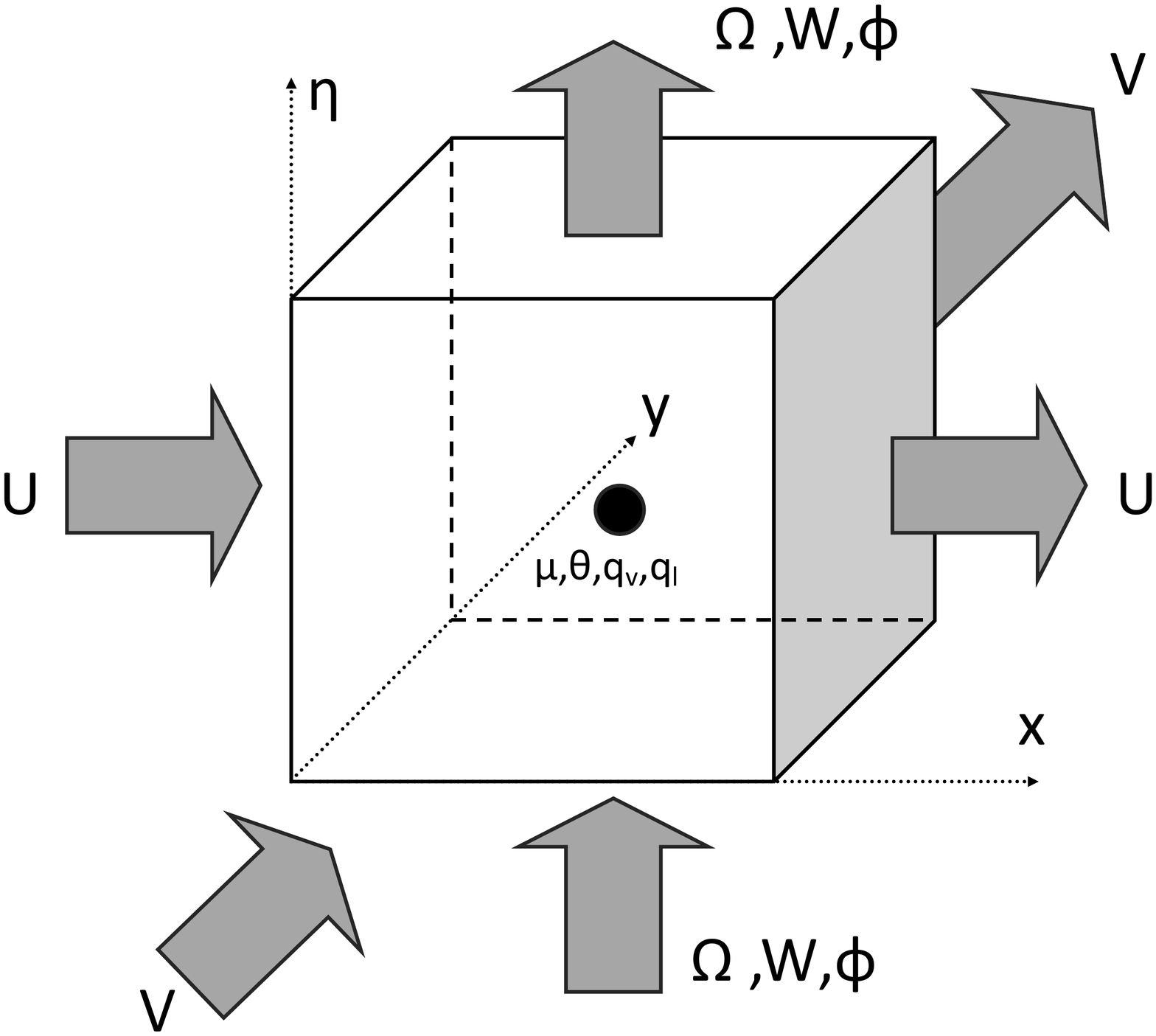}
\caption{Coordinates system of model's grid \cite{arwrf}}
\label{fig:grid}
\end {center}
\end{figure}

\paragraph{Grid and coordinates}
The atmospheric model operates on a logically quadrilateral 3D grid on the Earth's surface (Fig.\ref{fig:grid}). The vertical coordinate is the hybrid terrain-following mass coordinate $\eta$; defined as a combination between the terrain-following coordinate and a pure pressure coordinate.
Considering the hydrostatic pressure $p_d$ as an independent variable, ARW integrates the compressible, non hydrostatic Euler equations cast in the flux form using the following prognostic variables that have conservation properties (\cite{arwrf}, pp. 8-10): 

\vspace{-5pt}

\begin{itemize}
\itemsep-0.2em
    \item $\bm{v} = (u,v,w)$, where $\bm{v}$ are the covariant velocities in the horizontal and vertical directions, while $\omega=\partial_t \eta$ is the contravariant vertical velocity; 
    \item $\theta_m$ is the moist potential temperature; 
    \item $q_m= (q_v, q_c, q_r ...)$ represents the mixing ratios of moisture variables (water vapor, cloud water, rain water, ...). 
\end{itemize}
Since equations are formulated in flux form, the previous conserved prognostic variables have to be defined in flux form as well, leading to 
\begin{equation} \label{prognostic variables}
    \begin{cases}
    \bm{V} = \mu_d\bm{v} = (U,V,W) \\
    \Omega = \mu_d\omega \\
    \Theta_m = \mu_d\theta_m \\
    Q_m = \mu_dq_m \\
    \end{cases}
\end{equation}
where $\mu_d = \partial_\eta p_d$ is the vertical coordinate metric.
On the other hand, the geopotential $\phi = gz$, despite being a prognostic variable in the governing equations of the ARW, is not written in flux form because the quantity $\mu_d\phi$ is not conserved.

\vspace{-8pt}

\paragraph{Downscaling and nesting}
Even though the aim is to simulate microscale and mesoscale events, the synoptic and global scales cannot be neglected. However, a high resolution global model is computationally expensive. Through a downscaling it is possible to enhance the resolution starting from large scale information. WRF implementes dynamic downscaling through the nesting method that introduces an additional grid (\textit{child grid}) into the simulation with different granularity. Prognostic variable fields from the coarse-grid forecast are imposed as a boundary condition of the integration area. 

\subsection{\textbf{Fire module}}
The coupling of a mesoscale weather model (WRF-ARW model) with a 2D semi-empirical fire spread model permits the execution of a wildland fire simulation \cite{sfire}. SFIRE was developed based on the level set method i.e. an Eulerian approach. The reason why the fire model needs to be coupled with the atmospheric model is due to the significant influence of weather on fire behavior. In particular, the wind affects fire spread, while the fire causes changes in the atmosphere through the heat and vapor fluxes from burning fuel and evaporation of fuel moisture. For every WRF time step, one time step of the fire model is performed. The called fire model inputs the surface wind, which drives the fire, and outputs the heat flux into the atmosphere. The fire model has a high resolution and operates on a refined fire mesh contained in the inner grid of the atmospheric model.

\subsection{\textbf{Mathematical structure}} 

\paragraph{Atmospheric governing equations}
Using the variables defined in (\ref{prognostic variables}), the flux-form Euler equations, representing the core of the ARW, can be written as follows \cite{arwrf}
\vspace{1mm}

\noindent{\textit{Conservation of momentum}}
\begin{equation}
\begin{cases}
\partial_t U + (\nabla \cdot \bm{V}u) + \mu_d\alpha\partial_x p + (\frac{\alpha}{\alpha_d})\partial_\eta p\partial_x \phi = F_U \\
\partial_t V + (\nabla \cdot \bm{V}v) + \mu_d\alpha\partial_y p + (\frac{\alpha}{\alpha_d})\partial_\eta p\partial_y \phi = F_V \\
\partial_t W + (\nabla \cdot \bm{V}w) - g[(\frac{\alpha}{\alpha_d})\partial_\eta p - \mu_d]= F_w \\
\end{cases}
\label{eqn:Momentum}
\end{equation}
\textit{Conservation of heat}
\begin{equation}
    \partial_t \Theta_m + (\nabla \cdot \bm{V}\theta_m) = F_{\Theta_m}
    \label{eqn:Energy}
\end{equation}
\textit{Conservation of mass}
\begin{equation}
    \partial_t\mu_d + (\nabla \cdot \bm{V}) = 0
\label{eqn:Mass}
\end{equation}
\textit{Geopotential Material Derivative}
\begin{equation}
    \partial_t\phi + \mu_d^{-1} [(\bm{V} \cdot \nabla\phi) - gW] = 0
    \label{eqn:geopotential}
\end{equation}
\textit{Scalar moisture equations}
\begin{equation}
    \partial_t Q_m + (\nabla \cdot \bm{V}q_m) = F_{Q_m}
    \label{eqn:moisture}
\end{equation}
where $$\nabla \cdot \bm{V}a= \partial_x(Ua) + \partial_y (Va) + \partial_\eta (\Omega a)$$ and $$\bm{V} \cdot \nabla a= U\partial_xa + V\partial_ya + \Omega\partial_\eta a,$$ for a generic variable $a$.  

These equations are coupled with the \textit{diagnostic equation for dry hydrostatic pressure}
\begin{equation}
    \partial_{\eta} \phi = -\alpha_d\mu_d 
    \label{eqn:diagdry}
\end{equation}
and the \textit{ diagnostic relation for the full pressure} (dry air combined with water vapor)
\begin{equation}
    p=p_0 \Big (\frac{R_d\theta_m}{p_0\alpha_d}\Big)^{\gamma} 
\end{equation}

In Eq. \eqref{eqn:Momentum} - \eqref{eqn:moisture}, $\alpha_d = \frac{1}{p_d}$ is the inverse density of dry air, $\alpha= \alpha_d (1+q_v + q_c + q_r +...)^{-1}$, while $R_d$ is the gas constant for dry hair and $\gamma= \frac{c_p}{c_v}=1.4$ is the ratio of the heat capacities for dry air. The right hand side terms $F_U, F_V, F_W, F_{\Theta_m}$ contain the Coriolis and curvature terms along with mixing terms and physical forcings. 

Since the ARW solver works on a grid, a projection from the computational flat space to the physical spherical space is needed to interpret results: in particular, ARW implements the projection using map scale factors $m_x$ and $m_y$, defined as the ratio of the distance ($\Delta x, \Delta y)$ in computational space to the corresponding distance on the earth's surface: 
\begin{equation*}
    (m_x,m_y) = \frac{(\Delta x, \Delta y)}{\text{distance on the earth}}.
\end{equation*}
In practice, the ARW actually solves a perturbative form (\cite{arwrf}, pp.12) of the previous governing equations, where the momentum variables are redefined as follows

\begin{equation*}
    U= \frac{\mu_du}{m_y},
    V= \frac{\mu_dv}{m_x},
    W= \frac{\mu_dw}{m_y},
    \Omega= \frac{\mu_d\omega}{m_y}.\\
\end{equation*}

To solve the governing equations, ARW adopts a time-split integration scheme. In atmospheric model, however, slow and fast processes coexist and they have to be integrated differently: the former through a third-order Runge Kutta (RK3) time integration scheme; while the latter over a smaller RK3 time steps to guarantee numerical stability.   

\paragraph{Level set equation}
In SFIRE \cite{sfire}, the propagation of a fire burning in the area $\Omega=\Omega(t)$ in the horizontal $(x,y)$ plane on which the Earth is projected, is implemented by the level set method, which evolves a function $\psi = \psi(x,t)$, called the \textit{level set function}, such that the burning area at a time $t$ is $$ \Omega(t) = \{ x: \psi(x,t)\leq0\}$$ and the fireline $\Gamma(t)$, i.e. the boundary of the burning region $\Omega(t)$, is the level set 
\begin{equation}\label{fireline}
  \Gamma(t)= \{ x: \psi(x,t)=0\}.  
\end{equation}

The model adopts a semi-empirical approach, where the fire spread rate $S$ is computed from fuel properties, using the following modified Rothermel formula \cite{rothermel}: 
\begin{equation}\label{Rothermel}
    S= R_0 (1+ \phi_W + \phi_S).
\end{equation}
Here, $R_0$ represents the spread rate in the absence of wind, whereas $\phi_W$ and $\phi_S$ are respectively the wind factor and the slope factor. 
Given a point $ x \in \Omega(t)$, the time of ignition $t_i$ is defined as the time when the point $x \in \Gamma(t_i(x))$. 
The model assumes that at each location on the grid the fuel fraction, at the beginning $F=1$, decreases exponentially from the time of ignition $t_i$.
\begin{comment} 
as described by 
\begin{equation}\label{fuel fraction}
    F(x,t) = 
    \begin{cases}
    exp \Big (-\frac{(t-t_i)}{T_f} \Big ), & x \in \Omega, \\
    1, & \text{otherwise}, \\
    \end{cases}
\end{equation}
where $T_f$ is the fuel burn time, depending on fuel properties and taken to be independent from the wind speed and fuel moisture.  
$S$ is a function of the normal component of the wind factor $U$ and the terrain gradient $\nabla z$, meaning it is interpreted as a spread rate in the outside normal direction to the fire region. 
The spread rate $S$ in Eq. (\ref{Rothermel}) can also be rewritten as 
\begin{equation}\label{rothermel2}
    S= \max \big\{S_0, \; R_0 + c \min\{e, \max \{ 0, U\}\}^b + d\max\{0,tan\phi\}^2 \big \}, 
\end{equation}
where $S_0, \; R_0, \; b, \; c, \; d, \;e$ are fuel-dependent coefficients, stored for every grid point, that describes the spread within the fire. Thus the fire propagation speed can be input directly from past fire behavior for various kind of fuels.
\end{comment}

On the fireline, from Eq. (\ref{fireline}), the tangential component of the gradient $\nabla \psi$ is zero, meaning that the fire propagation speed is normal to the fireline $\Gamma$: the model postulates $S$ is a function of the normal component of the wind factor $U$ and the terrain gradient $\nabla z$ \cite{rothermel}. 

The evolution of the level set function is governed by the partial differential equation 
\begin{equation}\label{level set}
   \partial_t \psi + S ||\nabla \psi || = 0,
\end{equation}
called the \textit{level set equation}, which is solved numerically through discretization on the refined fire grid, adopting Heun's method, a second-order Runge Kutta method (\cite{sfire}, p.596). 
\subsection{\textbf{The new architecture}}

\begin{figure}[h]
\begin {center}
\includegraphics[width=0.518\textwidth]{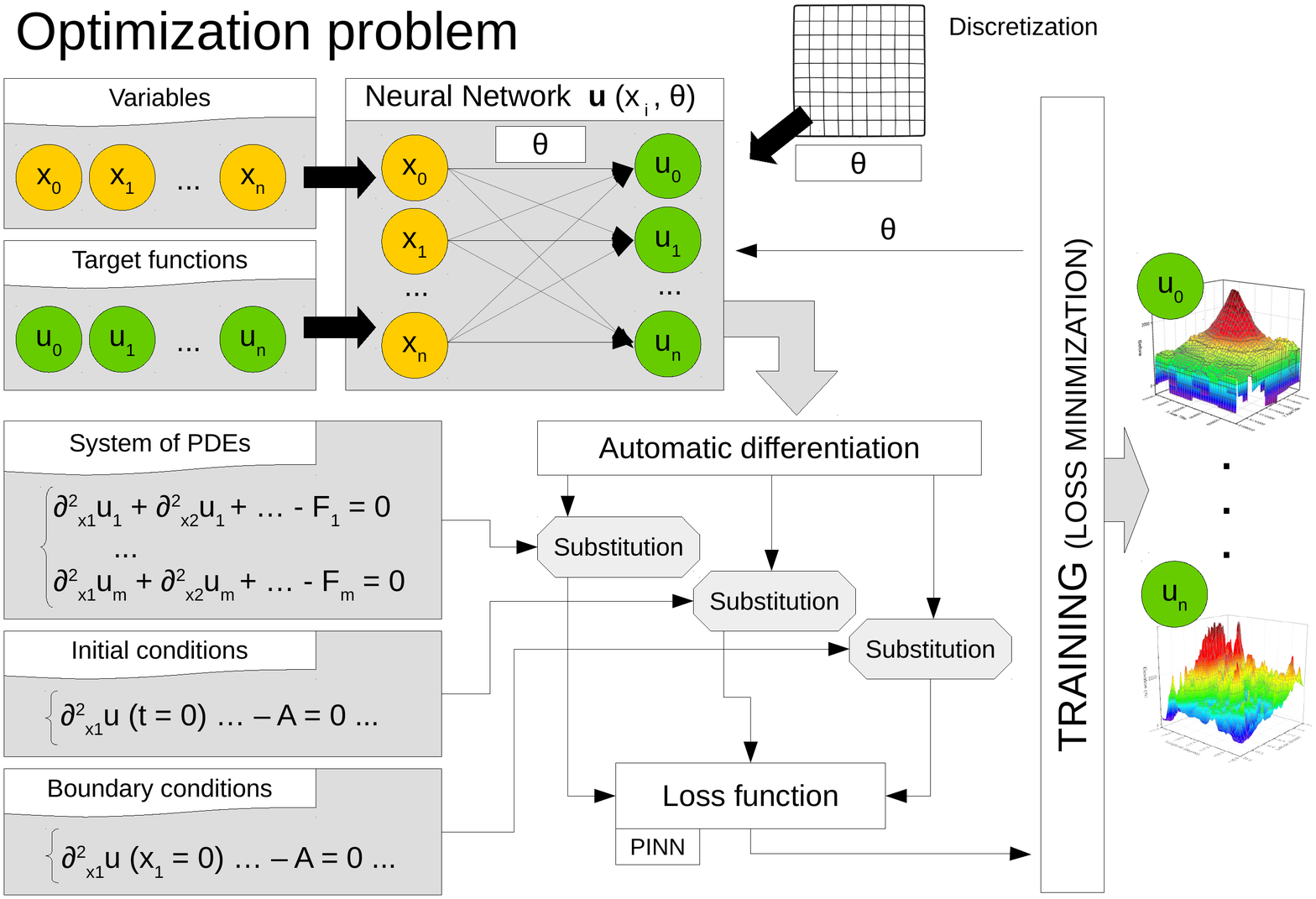}
\caption{Physics-informed neural network architecture for solving PDEs}
\label{fig:forcesneural}
\end {center}
\end{figure}

\vspace{-2pt}
We have replaced the original numerical solvers with a profoundly different approach, by taking advantage of the recently developed Julia packages \textit{NeuralPDE.jl} and \textit{DiffEqFlux.jl}, which support Physics Informed Neural Networks (PINNs) for automated PDE solving and Backwards Stochastic Differential Equation (BSDE) methods to deal with parabolic PDEs.
We employed the former method: PINNs are feed forward neural networks that are trained to solve supervised learning tasks while respecting any given law of physics described by general nonlinear partial differential equations. The resulting neural networks form a new class of universal function approximators that naturally encode any underlying physical laws as prior information \cite{kirill_nextjournal}.

The underlying architecture can be summarized by the following steps, with reference to Fig.\ref{fig:forcesneural}. Let's put ourselves in the most general case, where we want to find \textit{n} target functions $[u_1,...,u_n]$ which satisfy a system of \textit{n} PDEs. We first construct a surrogate solution to \textit{u(x)} as a neural network \textit{\^{u}(x;$\theta$)} with \textit{n} inputs and outputs and parameters $\theta$; since a neural network is mathematically a composite function, the derivatives of \textit{\^{u}} with respect to its input can be evaluated by applying the chain rule for differentiating compositions of functions using automatic differentiation (AD). Then we need to define a loss function (we used the $L^2$ norm) in order to calculate the discrepancy between $[u_1,...,u_n]$ and the outputs of the neural network: this is where we force our network to satisfy the physics imposed by the PDEs. Indeed, the loss is defined by substituting our neural network and its derivatives back into the equations we want to solve (where we brought all the terms on the left-hand side of the equal sign). The last step is to train the neural network to find the best parameters by minimizing the loss function with gradient-based optimizers, such as Adam or LBFGS. It is straightforward that minimizing the loss function means solving the equations, since we are approximating our target functions with \textit{\^{u}} better and better. \textbf{The key point is that it converts an integration problem into a mere minimization task and it does not need data to train, because it is trained on the PDEs themselves}. 

\subsection{\textbf{Model selection process that led us to this choice}}
Our aim from the beginning has been finding an ML alternative to standard numerical methods. However, we have explored different architectures within the field of Scientific Machine Learning.
We started our investigations using the \textit{DiffEqFlux.jl} library \cite{DiffEqFlux.jl}, which defines and solves neural ordinary differential equations (i.e. ODEs where a neural network defines its derivative function), which are implemented in the following way. First we need to get data from the numerical ODE solver, then define a neural network with a NeuralODE layer, which constructs a continuous-time recurrent neural network: at a high level this corresponds to solving the differential equation during the forward pass and using a second differential equation that propagates the derivatives of the loss function backwards in time. Now it's possible to train the model using an optimizer (ADAM or LBFGS) which minimizes the loss function.
We have consulted the library documentation and the GitHub resources to implement some toy models with this approach, such as the Lotka-Volterra equation, the 2D reaction diffusion equation, the 1D Fisher-KPP and the linear Burgers equation. Furthermore we tried to go beyond these implementations: we solved the reaction-diffusion equation with a Convolutional Neural Network (CNN) to investigate peculiarities of a CNN with respect to a common feed forward network in approximations of diffusive models.
When solving the Fisher-KPP we ran into a problem: since the neural ODE approximates the derivative of the solution, after it is trained we have not reached the desired solution yet; we still have to use a numerical solver. Following our initial purpose, we chose to add a Recurrent Neural Network on top of the NN + CNN architecture used in the first step.
Clearly this model showed some relevant disadvantages:
\begin{itemize}
\itemsep0em
    \item The predicted slices were very noisy, because the integrator would have required a greater amount of data for training;
    \item The integrator tended to overfit during the training, getting a model not applicable on other time scales and too \textit{problem-specific}.
\end{itemize}

This kind of issue made us lean towards the NeuralPDE PINNs approach, that is our ultimate architecture, clearly described in paragraph IV-D where the NN directly approximates the objective function, without needing further integrations.
Before attempting to model the complex PDEs inside WRF, we characterized the quality of the architecture by running it on the linearized Burger equation,
non linear Burger equation without viscosity, non linear Burger equation in the general form and the Poisson equation in 2 dimensions.

\subsection{\textbf{Implementation of the Level set}}
\vspace{-5pt}
The level-set differential equation is, as previously explained, the mathematical core for calculating the spread of the fire. The level set equation (that is a 2D surface embedded in a 3D euclidean space) is initially set as the distance of each point from the ignition's line. In our approach it is required that every quantity is provided either as a constant or as a function. Therefore we implemented the initial condition as a cone with elliptical section, because we had to deal with circular and linear ignition shapes (theoretically, any ignition shape can be given in input). The cone is then shifted down in order to make the contour level $z = 0$ resembling the desired initial fire line. Multiple ignition points are possible, but further refinements are needed to make this possibility fully functional.
The algorithm contains all the necessary expressions for the calculation of the fire spread rate, taking into account the values of the wind, the altimetric profile, the fuel map etc. These variables should be given in input as differentiable vector fields of space and time.
This can be considered both a limit and an advantage, in fact on one hand the discretized measurements must be fitted using approximating functions, but on the other hand this method is very computationally efficient. The possibility to give in input finite matrix of values is presented in "Future Work" and depends on the fact that some key libraries we use are still under development.
You can see in the appendix an example of the altimetric profile of the \textit{Isom Creek} fire location fitted using a 4th degree polynomial. 
The last step before the training of the model is a proportional scaling of the values to a domain with size in the order of $10^{1}$ and the definition of a discretization that is used only to run the training algorithm. The outputs will be continuous functions. The discretization to this interval prevents some numerical instabilities and spikes that are linked to the early stage of development of some Julia libraries in use. The loss function is made of two parts: the former minimizing the $L^2$ error referred to the PDE definition and the latter minimizing the $L^2$ distance between the boundary conditions and the solution evaluated at the boundaries.  At this point the training algorithm is instantiated and launched. The minimization of the loss functions is the process that actually solves the PDE and constitutes the main load for the CPU. It can be easily parallelized and thus accelerated using GPUs.
When the training is completed the prediction undergoes a new proportional scaling that gives back the original domain shape.

\subsection{\textbf{Implementation of the Euler system}}
Writing and solving the 7-equation Euler system in Julia was really challenging, in fact at the moment we are not aware of any publications where these techniques are yet applied to PDE systems of such complexity. The first problem was to figure out which were the independent and dependent variables.: it might seem rather simple, but climate models adopt different conventions depending on the circumstances (e.g., they generally use pressure coordinates instead of height coordinates). That is why we presented various Julia files, both with six and seven target functions, for reasons we cannot explain here for brevity. Since the chain rule for derivatives hasn't been implemented into \textit{ModelingToolkit.jl} yet, one of the most demanding operations was writing all the derivatives involved by symbolic calculation.
Then we had to choose the most suitable training strategy, which turned out to be \textit{QuadratureTraining}. We have used parabolic initial conditions for a first evaluation, but we want to investigate them more in depth, since the study of boundary conditions often deserves a paper on its own. Unfortunately, the \textit{NeuralPDE} library is still unable to treat this kind of problem with stability, and often incurs errors due to internal divergence calculations. Despite this, we have been able to obtain convergence of the loss function, although it is not enough to present valid results. We contacted the authors of these libraries, that are still under development, and we are looking forward to contributing.

\subsection{\textbf{Optimization and coding}}
For the implementation of the model, we chose to rely on the Julia programming language. Julia, despite being a very young language with a huge amount of bugs and improvements still to be implemented, compared to the more diffused Python, offers better performance since it is not an interpreted language providing Just in Time (JIT) compilation with different optimization levels \cite{julia_perf}. Moreover, Julia is particularly well suited for numerical computation and for the solution of complex physical models, both considering the syntax and the presence of some specific libraries such as \textit{DifferentialEquations.jl} and \textit{ModellingToolkit.jl}.  

The model was implemented using the low-level interface of the \textit{NeuralPDE.jl} library which contains the necessary methods for the generation of the training datasets and of the loss functions starting from the explicit form of the equations and the boundary conditions. This library allows some very high level macros for the declaration of derivatives, functions and parameters and allows you to choose the differentiation techniques (numerical or automatic) and the optimization engine to be used at a later time. The library relies in turn on \textit{GalacticOptim.jl}, which provides the methods for minimizing the loss, and on \textit{ModelingToolkit.jl}, which provides the constructors for the derivatives and on \textit{Flux.jl}, which is the library that implements the constructors of neural networks with the respective backpropagation and prediction functions. \\

\vspace{-6mm}

\subsection{\textbf{Synthetic datasets - ideal simulations}}
The WRF source code includes some idealized simulations that are used as a playground for the model: these simulations are composed of an namelist.input file and an input\_sounding file. The former contains both data about the physics models that should be used, as well as terrain information, to generate the surface height mesh; the other is a standard way of storing weather information, containing potential temperature, water vapor mixing ratio, and U and V wind speed components values at various pressure values.

We decided that it was best to start with an ideal simulation to be able to control the response of the model, so we modified the already present \textit{two\_fires} test to fit our needs, reducing the number of fires to one and increasing the simulation time to 60 minutes.

The terrain generated is completely flat, and the entire domain is covered in tall grass (Anderson fuel category 3 \cite{anderson}). Since the wind is entirely along the V component, we expect the fire to expand upwards. 

\subsection{\textbf{Real datasets - real simulations}}
In order to run a real data wildfire simulation in the WRF model, static terrestrial data, atmospheric data and a map of the 13 Anderson's fuel category of the desires domain are required. The geographic global static data containing the mandatory fields requested are provided by the  National Center for Atmospheric Research (NCAR) website \cite{UCAR2}; whereas an Anderson 13 fuel categories \cite{anderson} map and a high-resolution topographic data (about 5 meters resolution) are provided by the Landfire \cite{landfire2} site. Lastly, for our simulation we adopted historical meteorological fields from a Global Forecast System (GFS) reanalysis, with a resolution of 0.25 degree and provided with a 6-hour time step. The latter are accessible in Grib2 format on the Research Data Archive RDA of NCAR website \cite{GFS_site}.

\begin{figure}[ht!]%[!ht]
\begin {center}
\includegraphics[width=0.4\textwidth]{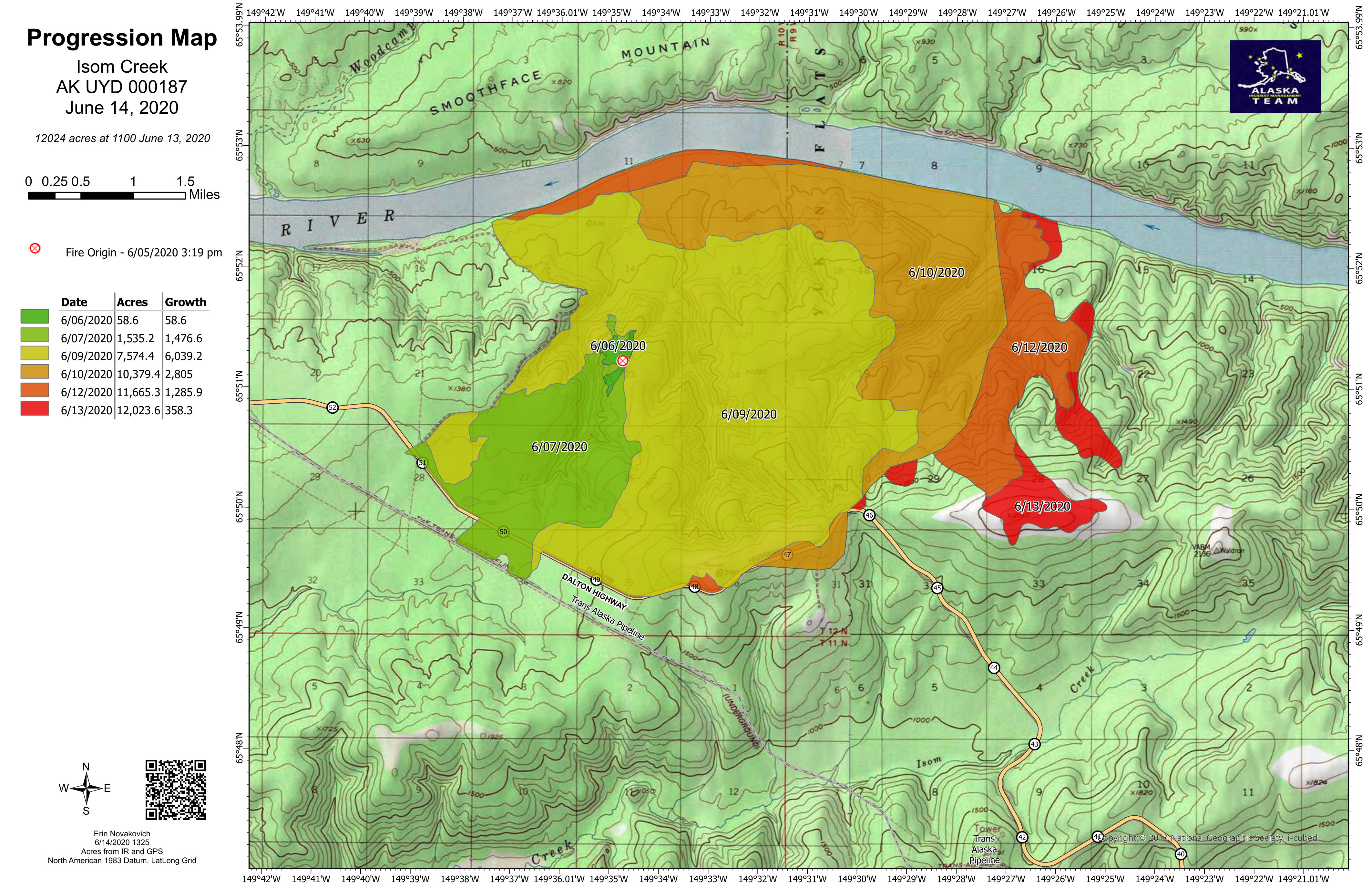}
\caption{Isom Creek Fire progression map}
\label{fig:progression_map}
\end {center}
\end{figure}

All the datasets were appropriately pre-processed by the WRF Pre-Processing System (WPS), which allows to define the simulation domain. In addition, through a horizontal interpolation of datasets, WPS provides the initial and boundary conditions that will be used by the WRF model. 
For this work we carried out a simulation of the Isom Creek fire, Alaska (USA), \cite{isom_creek_info} where the ignition happened on 05 June 2020 at 03:15 PM and the progression lasted until 13 June, reaching an extension area of 12,180 acres (Fig.\ref{fig:progression_map}). 

\begin{figure}[ht!]%[!ht]
\begin {center}
\includegraphics[width=0.45\textwidth]{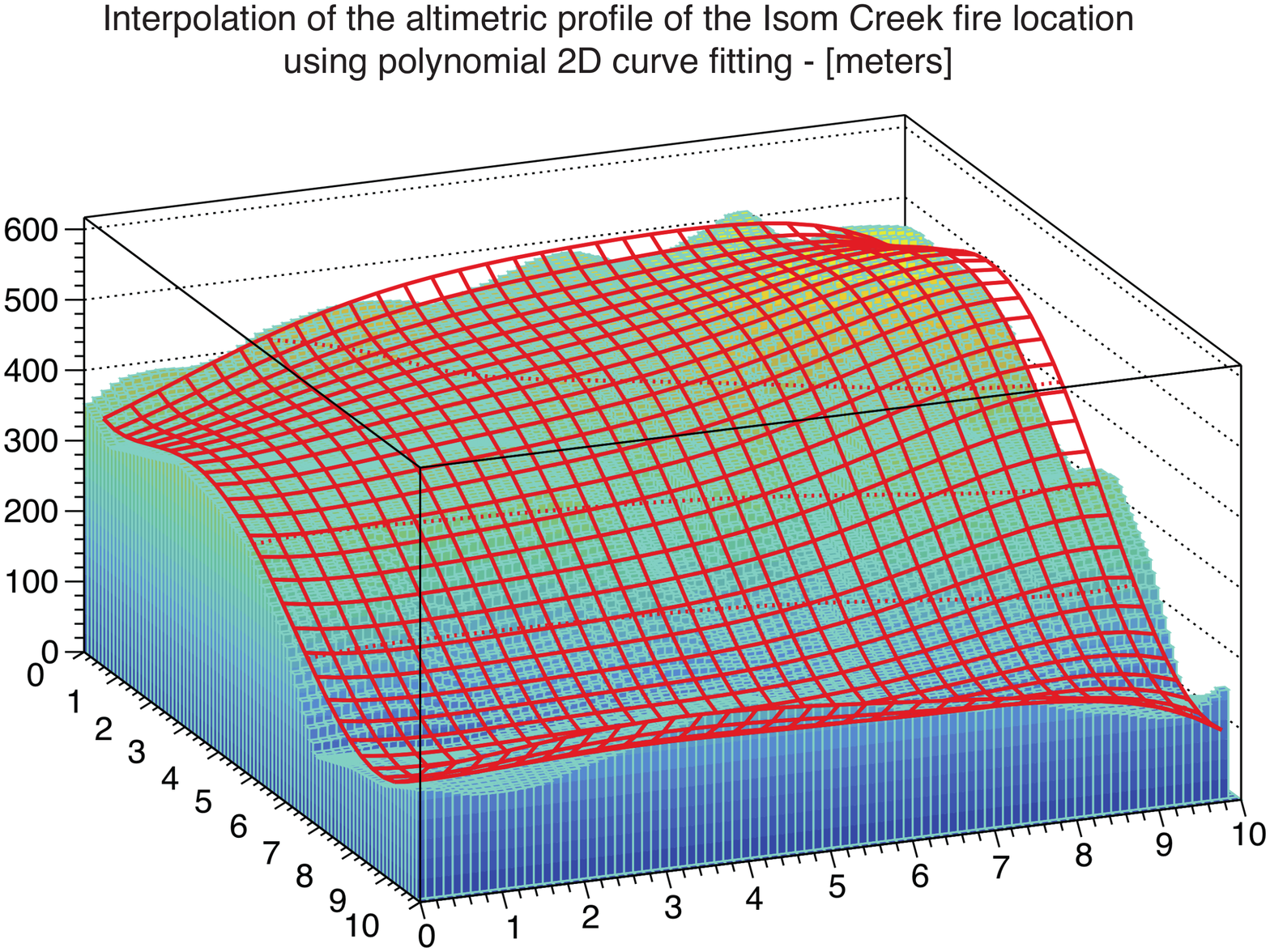}
\caption{Altimetric profile interpolation of the Isom Creek fire region.}
\label{fig:interpolation}
\end {center}
\end{figure}

\begin{figure}[ht!]%[!ht]
\begin {center}
\includegraphics[width=0.5\textwidth]{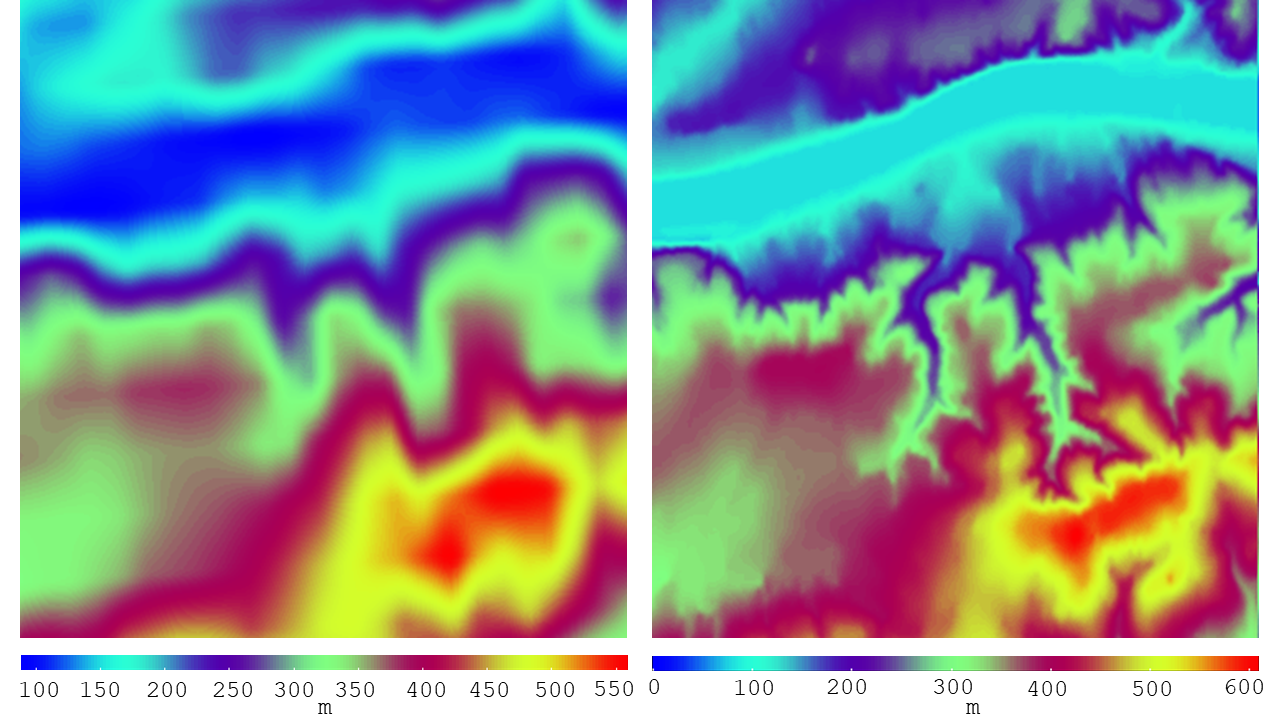}
\caption{Comparison between global topological data and high resolution data after \textit{geogrid.exe} program run}
\label{fig:HGT_ZSF}
\end {center}
\end{figure}

\begin{figure}[ht!]%[!ht]
\begin {center}
\includegraphics[width=0.3\textwidth]{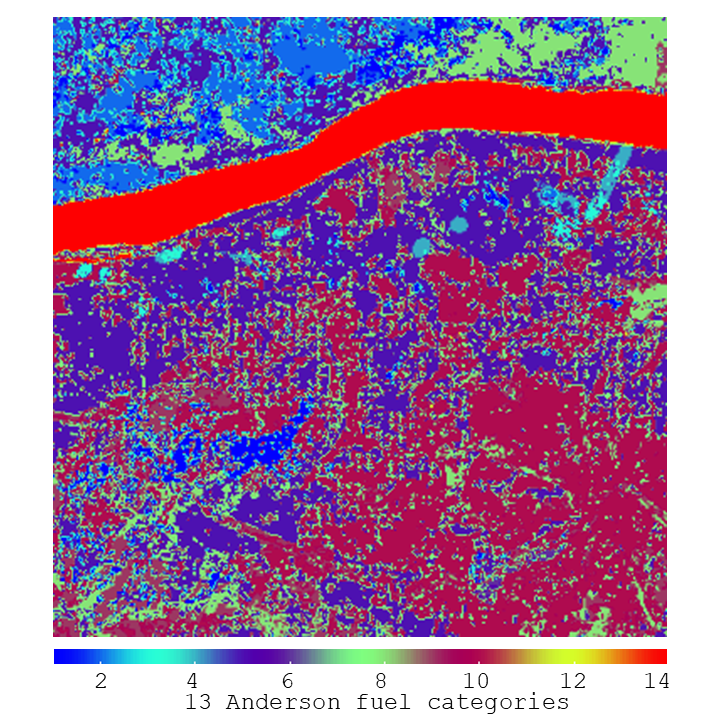}
\caption{Overlap of the 13 Anderson's categories map on the domain of simulation after the \textit{geogrid.exe} program run}
\label{fig:NFUEL_CAT}
\end {center}
\end{figure}

Therefore, the center of the simulation domain has been fixed at the coordinates of the Isom Creek fire ignition point (65.854; -149.579). The chosen atmospheric domain is a mesh of 97x97 grids, each with a side of 100 m, while the 2D mesh of the fire domain is 20 times finer, therefore with a resolution of 5 meters. The extrapolation of the domain from the land-surface datasets takes place by means of the \textit{geogrid.exe} program in WPS, and at this point occurs also the appropriate overlap between all the input geographical data Fig.\ref{fig:HGT_ZSF} and Fig.\ref{fig:NFUEL_CAT}

By briefly analyzing these datasets we found a few interesting facts. The most noticeable aspect is the fuel distribution: the most abundant fuel is timber (category 10), followed by brush (category 5) and closed timber litter (category 8). This fuel composition is wide-spread in this area and due to the presence of timber it can lead to potential fire control difficulties \cite{anderson}. The presence of a river in the top part of the domain suggests that the fire will not spread here.

The mountainous topography of the area will also affect the spread of the fire, as the terrain gradient is part of the calculation for the fire spread rate.

\textit{ungrib.exe} deals with the pre-processing of global atmospheric dynamical data. Finally, the \textit{metgrid.exe} program is responsible of the combination of meteorological data to the simulation domain.

Some attempts of simulations with a second finer domain, inner to the domain described above, have been made. The side of the inner domain is a third of the outer domain, and the number of cells is also 97, therefore we obtained a domain of about 3 km with a resolution three times higher. From these tests we obtained relatively valuable simulations, bearing in mind that for larger simulations researchers set the beginning of the simulation at least 12 hours before of the period of interest. However, we preferred to come back to a single domain simulation in order to speed up the run, paying the price in terms of a meteorological simulation affected by boundary effects. Our ultimate Isom Creek fire simulation ran for 18 hours, starting from the 5th June at 12:00 PM.

\subsection{\textbf{Ideal simulation: "One fire"}}
\begin{figure}[ht!]%[!ht]
\begin {center}
\includegraphics[width=0.45\textwidth]{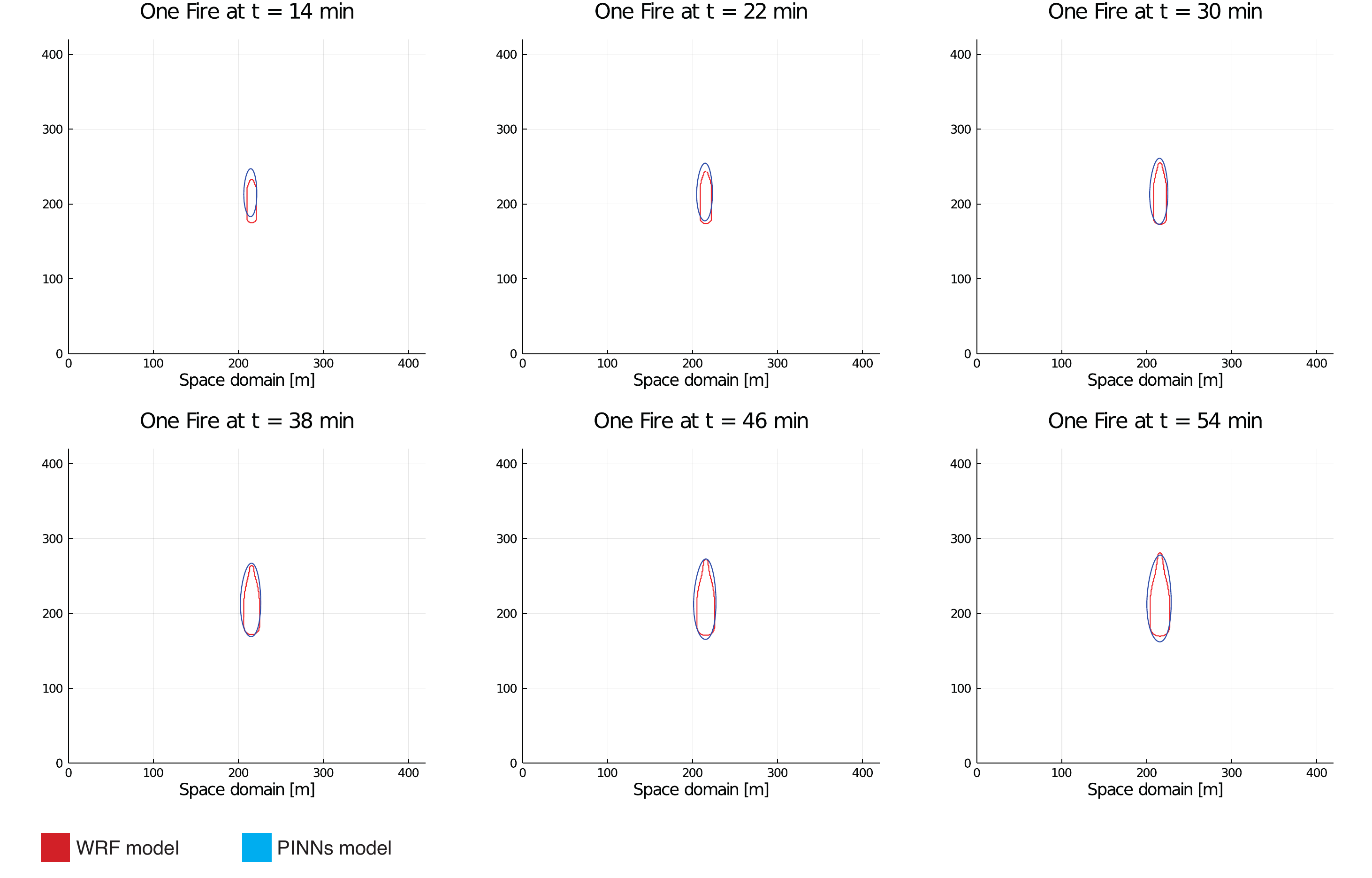}
\caption{Comparison between the WRF and PINNs outputs at different time frames}
\label{fig:one_fire_sequence}
\end {center}
\end{figure}

For clarity, we renamed the \textit{Two Fires} ideal simulation as \textit{One Fire}, after having removed one ignition point. It is a simple case that demonstrates that our model works well in case of fires that spread in flat areas where the fuel doesn't change. It also evolves according to the wind direction, represented by a vector with two components. The results at different time frames are represented in Fig.\ref{fig:one_fire_sequence}. %
The details about the architecture employed and the training are listed in TABLE \ref{fig:specs_one_fire}.

\vspace{2pt}

\begingroup
\setlength{\tabcolsep}{9pt} 
\renewcommand{\arraystretch}{1.2}
\begin{table}[h]
    \begin{tabular}{  l  p{3.0cm}  }
        \toprule
\textbf{Optimization Algorithm}      
& ADAM   
 \\\hline
 
\textbf{Iterations}
& 4800      
 \\\hline
 
\textbf{Final Objective Value}
& 6.39 e-8 
 \\\hline
 
 \textbf{Training Strategy}
& QuadratureTraining()  
 \\\hline
 
\textbf{Domains}
& $t\in[0,10]$, $x\in[0,10]$, $y\in[0,10]$
 \\\hline
  
\textbf{Training Mesh Size}
& $[dt,dx,dy] = [0.17,0.02,0.02]$  
 \\\hline
 
\textbf{Boundary Condition}
& $u(0,x,y,\theta) = ((5 (x - 0.3))^2 + (0.15 y)^2)^{\frac{1}{2}} - 0.2$ 
\\\hline

\textbf{Neural Network dimensions}
& $3>16>1$
\\\hline

\textbf{Training Time}
& 647 s
\\\hline
\bottomrule
    \end{tabular}
    \caption{Technical Specs of the model - One Fire} 
    \label{fig:specs_one_fire}
\end{table}
\endgroup

\begin{figure}[ht!]%[!ht]
\begin {center}
\includegraphics[width=0.4\textwidth]{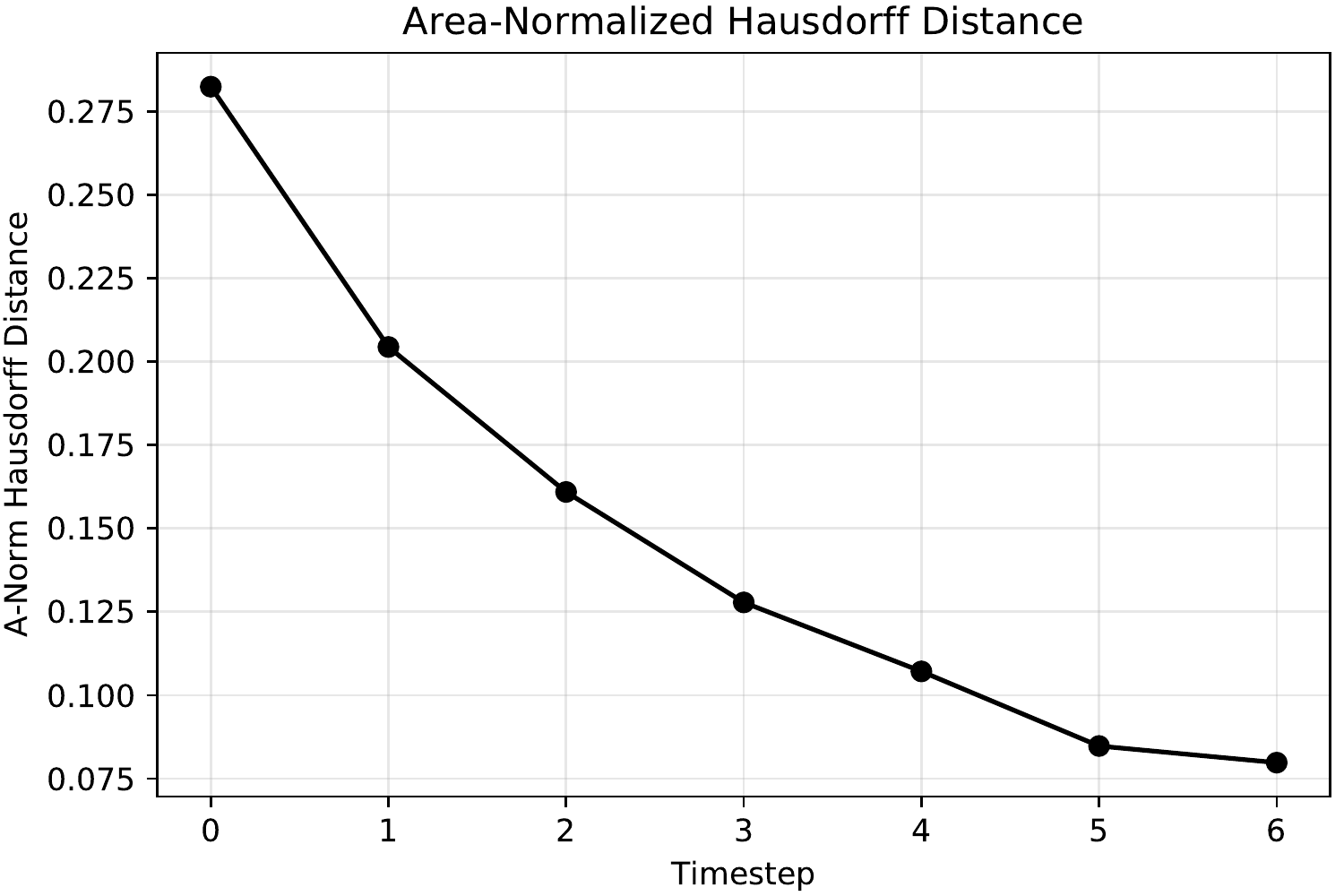}
\caption{Hausdorff distance as a measure of the error - One Fire.}
\label{fig:hausdorff_one_fire}
\end {center}
\end{figure}

In order to provide a quantitative measure of the error between the outputs we decided to use the Hausdorff distance \cite{hausdorff},
\begin{displaymath}
    d_{\mathrm H}(X,Y) = \max\left\{\,\sup_{x \in X} \inf_{y \in Y} d(x,y),\, \sup_{y \in Y} \inf_{x \in X} d(x,y)\,\right\},
\end{displaymath}
where $X$ and $Y$ are the firelines to be compared. This distance is further normalized on the area of the curves that represent the fireline, as Fig.\ref{fig:hausdorff_one_fire} shows. The fact that the error decreases tells us that as the area grows with time, the difference between the two regions doesn't increase, meaning that they evolve in synchrony and with the same shape.

\vspace{-7pt}

\subsection{\textbf{Simulation of a real wildfire: Isom Creek, Alaska, 6/5/2020}}

\begin{figure}[ht!]%[!ht]
\begin {center}
\includegraphics[width=0.45\textwidth]{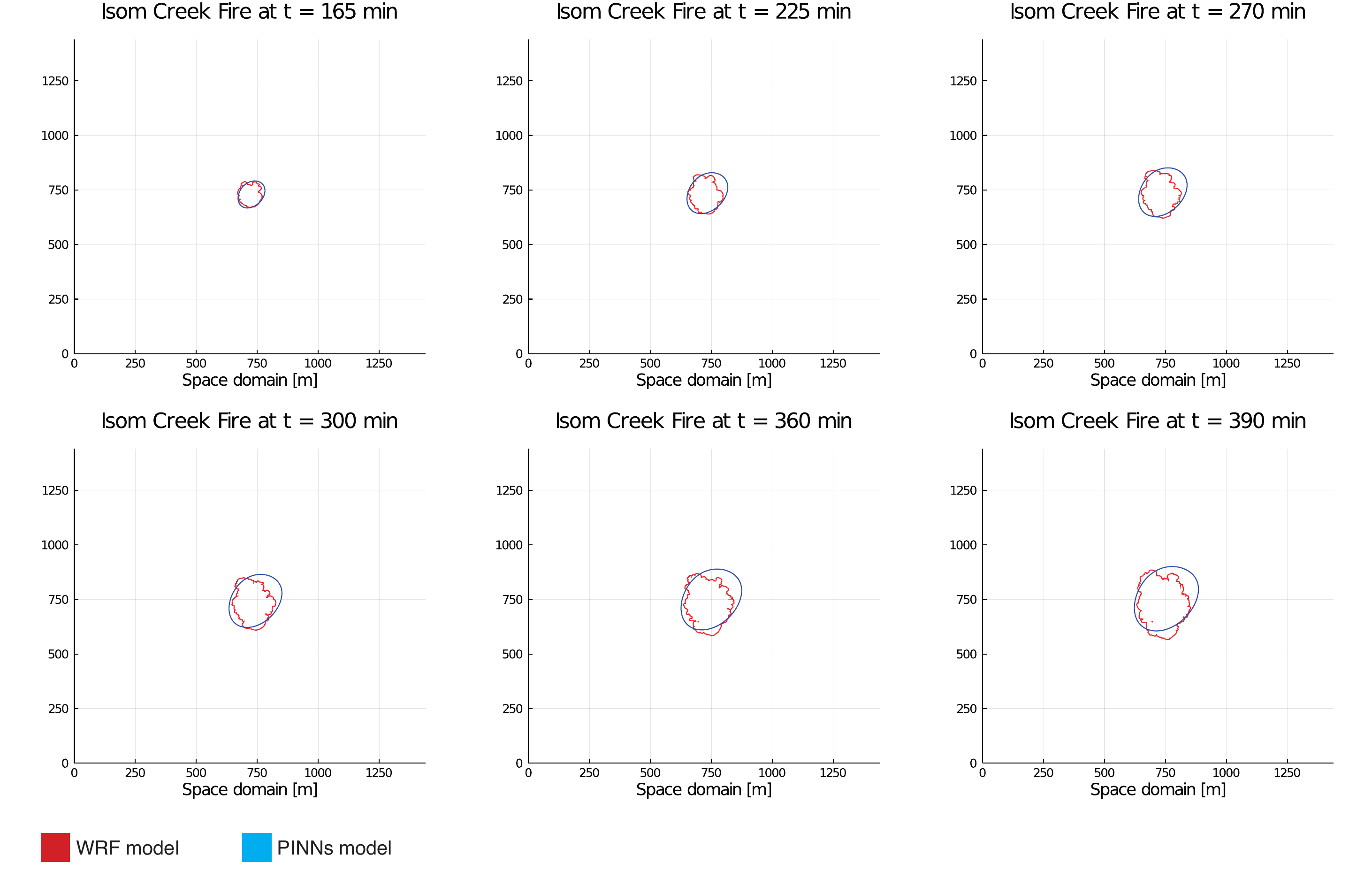}
\caption{Comparison between the WRF and PINNs outputs at different time frames}
\label{fig:isom_sequence}
\end {center}
\end{figure}

We have been able to simulate the Isom Creek fire both with WRF and with our implementation. Although it is not possible to perform a precise quantitative comparison between WRF and the real data, or between our result and the real data (due to a lack of temporal precision of the available data at \cite{isom_creek_map}), it is clear that the WRF output simulates the wildfire accurately, at least visually (see Fig.\ref{fig:progression_map}).
The result obtained with the PINN architecture is shown in Fig.\ref{fig:isom_sequence}.
It is clear that our implementation hasn't been able to capture the shape of the fireline with the same accuracy. This will be pointed out below as a temporary limitation to our model: at the moment it is not possible to assign a different fuel category to each mesh point (i.e. using a matrix to build a function), so the fireline will always have a symmetrical and smooth perimeter.
The specifics of the architecture follow.

\vspace{2pt}

\begingroup
\setlength{\tabcolsep}{9pt} % Default value: 6pt
\renewcommand{\arraystretch}{1.2}
\begin{table}[h]
    \begin{tabular}{  l  p{3.0cm}  }
        \toprule
\textbf{Optimization Algorithm}      
& ADAM   
 \\\hline
 
\textbf{Iterations}
& 3000     
 \\\hline
 
\textbf{Final Objective Value}
& 2.14 e-6
 \\\hline
 
 \textbf{Training Strategy}
& QuadratureTraining()  
 \\\hline
 
\textbf{Domains}
& $t\in[0,10]$, $x\in[0,10]$, $y\in[0,10]$
 \\\hline
  
\textbf{Training Mesh Size}
& $[dt,dx,dy] = [0.17,0.02,0.02]$  
 \\\hline
 
\textbf{Boundary Condition}
& $u(0,x,y)=((x-5)^2+(0.7 (y-5))^2)^{\frac{1}{2}}-0.2$
\\\hline

\textbf{Neural Network dimensions}
& $3>16>1$
\\\hline

\textbf{Training Time}
& 270 s
\\\hline
\bottomrule
    \end{tabular}
    \caption{Technical Specs of the model} 
    \label{fig:specs_isom}
\end{table}
\endgroup

\begin{figure}[ht!]%[!ht]
\begin {center}
\includegraphics[width=0.4\textwidth]{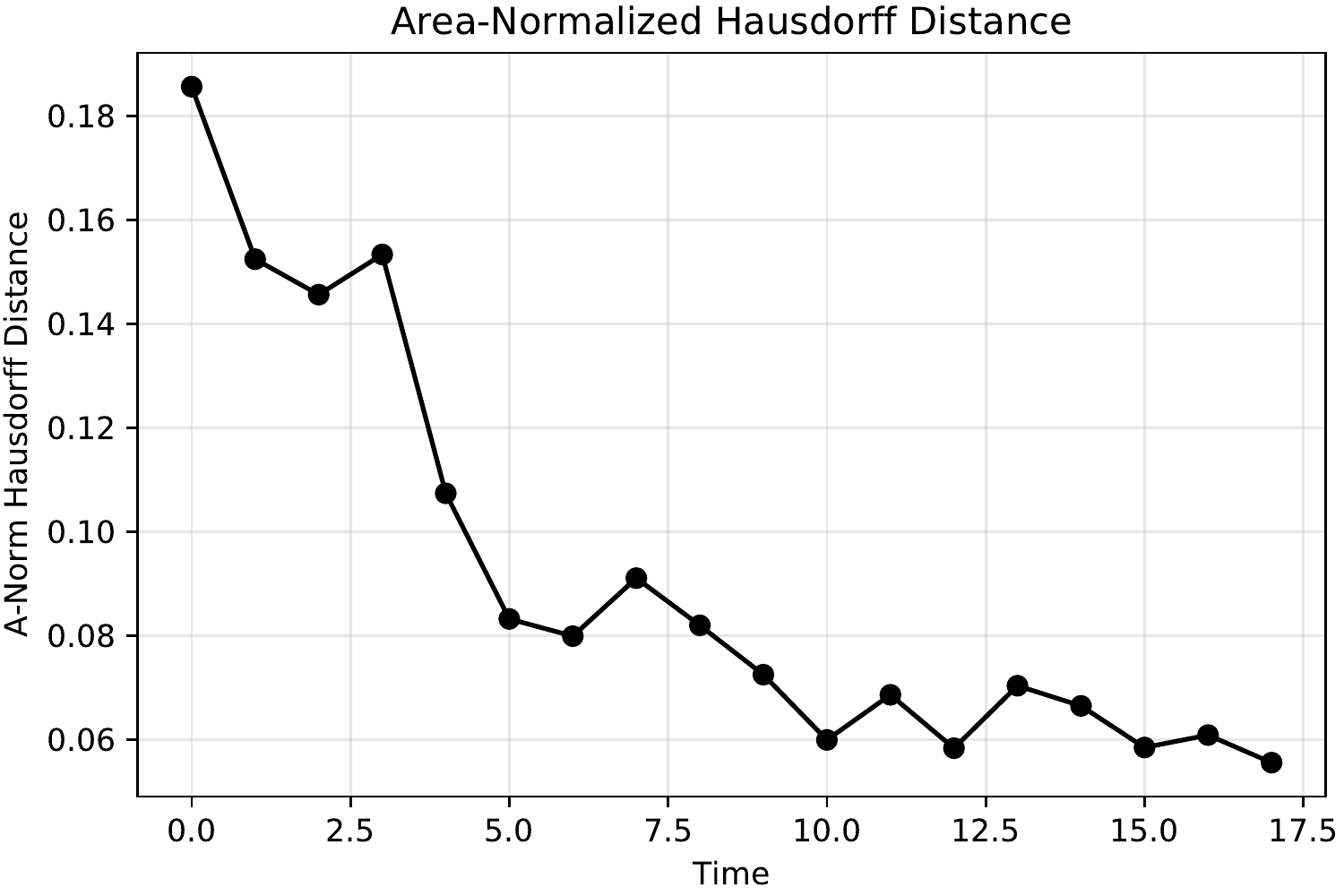}
\caption{Hausdorff distance as a measure of the error - Isom Creek.}
\label{fig:hausdorff_isom}
\end {center}
\end{figure}

The error can be quantitatively measured with the Hausdorff distance as before, with the outcome plotted in Fig.\ref{fig:hausdorff_isom}.  

The considerations outlined in the previous paragraph remain valid here (see \ref{appendix} for additional comparison plots).

\section{\textbf{DISCUSSION}}\label{discussion}

\subsection{\textbf{Discussion of the experimental results}}
The results exposed above show the feasibility of utilizing the Julia \textit{NeuralPDE} library to model a wildfire related differential equation, i.e. the level set equation. As disclosed above, we have also obtained some preliminary but unstable results regarding the atmospheric coupling of the model by attempting to solve the Flux-Form Euler system equation.
The outputs obtained with our approach don't strictly follow the jagged firelines of WRF: that's not an intrinsic limit of the architecture, which doesn't lack variance even if it is rather simple; the reason is, as explained above, that the inputs (fuel and wind in particular) we gave to the NN are not varied enough, due to limitations of the library.
In spite of this we have shown the ability of our implementation to reproduce the desired time evolution of the fireline as given by WRF.

\subsection{\textbf{Advantages of this architecture}}\label{advantages}
The various advantages of this approach are:

\vspace{-5pt}
\begin{itemize}
\itemsep0em
\item We can get a rough estimate of the speed up our model provides - compared to numerical solvers - by noting that the train time is of some hundreds of seconds, while the WRF run time is more than an hour. This point has to be further studied: on one hand, we used a simpler set of inputs for the reasons we will explain later; on the other hand we worked on one of the eight cores at our disposal.
\item While numerical methods only solve on the initial domain range, this architecture learns the neural network parameters that can be used to \textbf{extend the predicted solution outside of the training domains} (with a decrease in accuracy if the extrapolation is extended too much). This can be used for "forensic investigations": it is possible to reconstruct the ignition point and the evolution of the fireline from the final state of the wildfire.
\item It is possible to interpolate the solution on a continuous mesh.
\item Modifying the equations of the model is as easy as changing a few lines of code, instead of re-implementing the discretization and perturbative form of the equations, allowing for faster investigation of new models.
\item Given the possibility to use several CPU cores, the speed could increase even more and this approach could be run iteratively in order to simulate the outcome of different \textit{containment scenarios} and choose the best among them.
\item This architecture doesn't need neural networks with a high number of degrees of freedom to be accurate, reducing the overall computational cost and increasing performance. In quantitative terms, this means not exceeding some tens of neurons on a single hidden layer. 
\item As a consequence, the model is more interpretable (it is less similar to a complete \textit{black-box}) and its users can run the resulting simulator on standard PCs without needing high performance machines.
\item PINNs can be used to extrapolate physical laws once they have been interpolated, hence providing a tool to refine theoretical models. However, this requires a greater amount of input data.
\end{itemize}

\subsection{\textbf{Limitations}}
Let's now analyze the limits of our approach:
\vspace{-5pt}

\begin{itemize}
\itemsep0em
\item At the moment, it cannot be run on GPUs nor multiple CPUs because the library we used is not fully developed yet. We contacted the developers and they are working on it.
\item The library doesn't support the usage of matrices as functions of their indices. This means it is not possible to use a variable fuel map, fundamental in order to capture the evolution of the fire with higher accuracy. We opened an issue about this \cite{issue177}.
\item Our approach suffers from numerical instability depending on the training strategy adopted and on the number of neurons and layers.
\item The model only works on relatively small domains and is unstable for larger ones. Therefore we had to scale down all the quantities involved and introduce scale factors on the fire spread rate in order to make it compatible to the evolution of the output of WRF.
\item We still have to figure out how to predict the spread of a wildfire with more than one ignition point.
\item The architecture has not yet been extended to Convolutional Neural Networks, Recurrent Neural Networks or others.
\end{itemize}

\section{\textbf{FUTURE WORK}}
\begin{figure}[ht!]%[!ht]
\begin {center}
\includegraphics[width=0.4\textwidth]{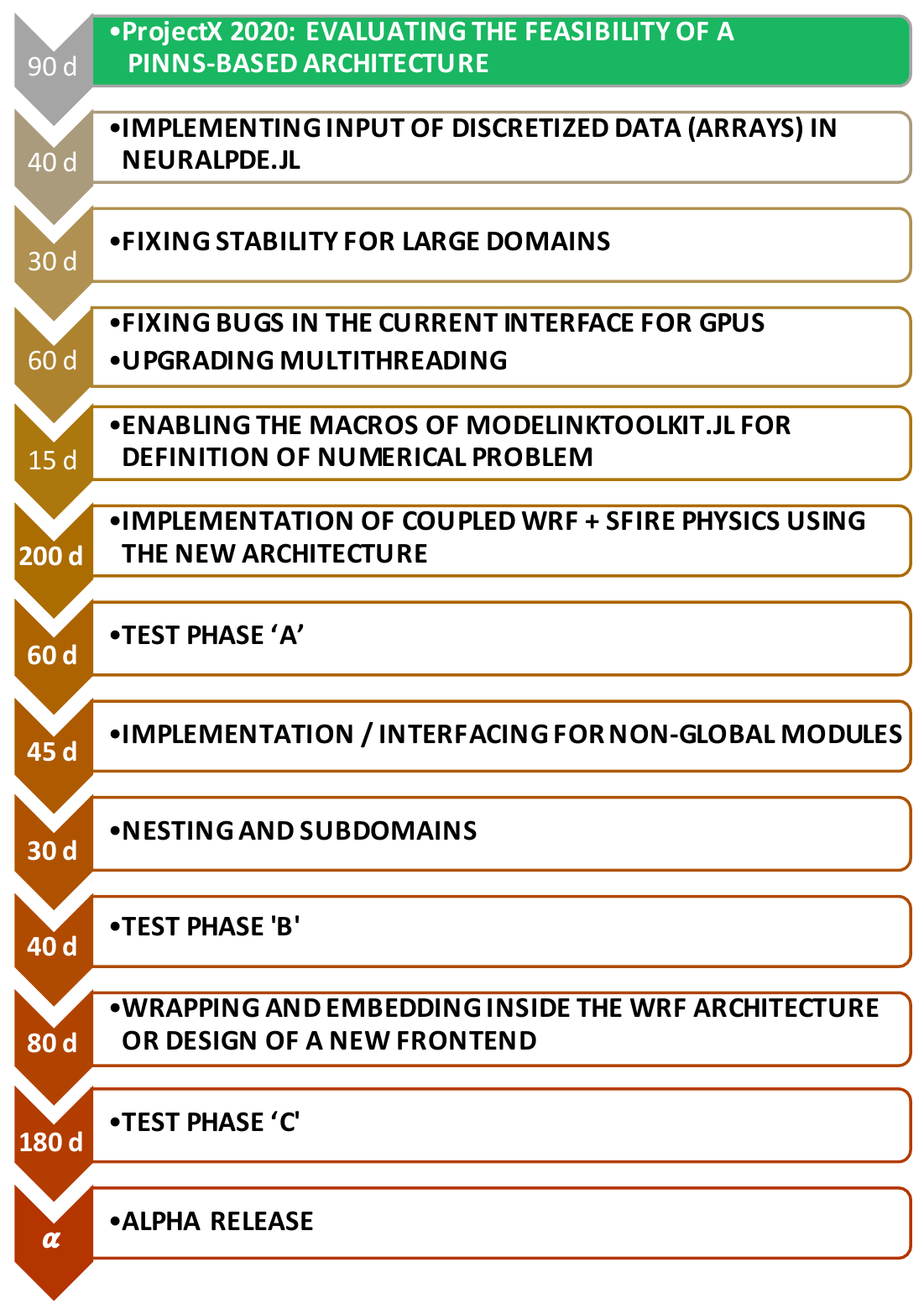}
\caption{Future prospect of our work}
\label{fig:future_work}
\end {center}
\end{figure}

This research is the first step towards a concrete implementation of PINNs-based wildfire simulation. We have designed a chart (Fig.\ref{fig:future_work}) that shows the next improvements to be made in order to achieve such a ambitious goal.  

\section{\textbf{HARDWARE SUPPORT AND CODE}}
The implementation described in the previous paragraph is based on the ML framework \textit{Flux.jl} that is agnostic with respect to the array type. This allows to also pass to this engine GPU arrays (\textit{CUDA arrays}) that are allocated and computed on GPUs (General Purpose GPUs). For the development we used a workstation with the following technical specifications:
\vspace{-5pt}

\begin{itemize}
\itemsep0em
    \item CPU: Intel Core i7 10700K 8 cores / 16 threads
    \item RAM: 64 GB DDR4 unbuffered
    \item GPU: Nvidia 2080 Ti 11 GB GDDR6
\end{itemize}

\vspace{-5pt}
Unfortunately, the GPU interface of \textit{NeuralPDE.jl} is not mature yet.

All the code produced is published on a public repository available at this link: \\ \url{https://github.com/MachineLearningJournalClub/MLJC-UniTo-ProjectX-2020-public}.\\
The output of the WRF simulations can be found in the following Google Drive folder: \\  \url{https://drive.google.com/drive/folders/1wUCKUyVwC0Pf-e9WlLiqOxRLF0or2D0U?usp=sharing} \\ or \url{https://tinyurl.com/mljc-unito-px2020}.

\section{\textbf{SOCIETAL CONSEQUENCES}}\label{societal consequences}
\subsection{\textbf{The impact of real time simulations for containment}}
As previously mentioned, one of the most important aspects of implementing a fast simulator is the possibility of employing it in wildfire containment, as firefighters could simulate the possible outcomes for different strategies, therefore applying the best one. At the moment WRF-SFIRE is not fast enough to be used this way, but the ML approach shows promising results for such an application. As of now, firefighters still struggle to predict changes in fire spread, as we observed even in the Isom Creek fire reports \cite{isom_creek_spread}, which lead to the destruction of over 12,000 acres of forest, a huge economic cost, and the damage done to an entire ecosystem.
The usage of simulators with the characteristics we outlined would have the ethical value of saving lives, on top of saving huge amounts of money by reducing the targeted area.

\subsection{\textbf{Simulation of fire propagation and climate change}}
A fast and reliable wildfire simulation can also be used to simulate how wildfires might develop and worsen in different climatic conditions, for example by reproducing the same fire in various scenarios. This can be done by changing some of the input meteorological data, such as the temperature and the radiative forcing (RF). Using predictions of how climate might change due to global warming, the model could prove the impact that this crisis might have on wildfires.

\subsection{\textbf{Forensic investigation}}\label{Forensic_investigation}
\begin{figure}[ht!]%[!ht]
\begin {center}
\includegraphics[width=0.3\textwidth]{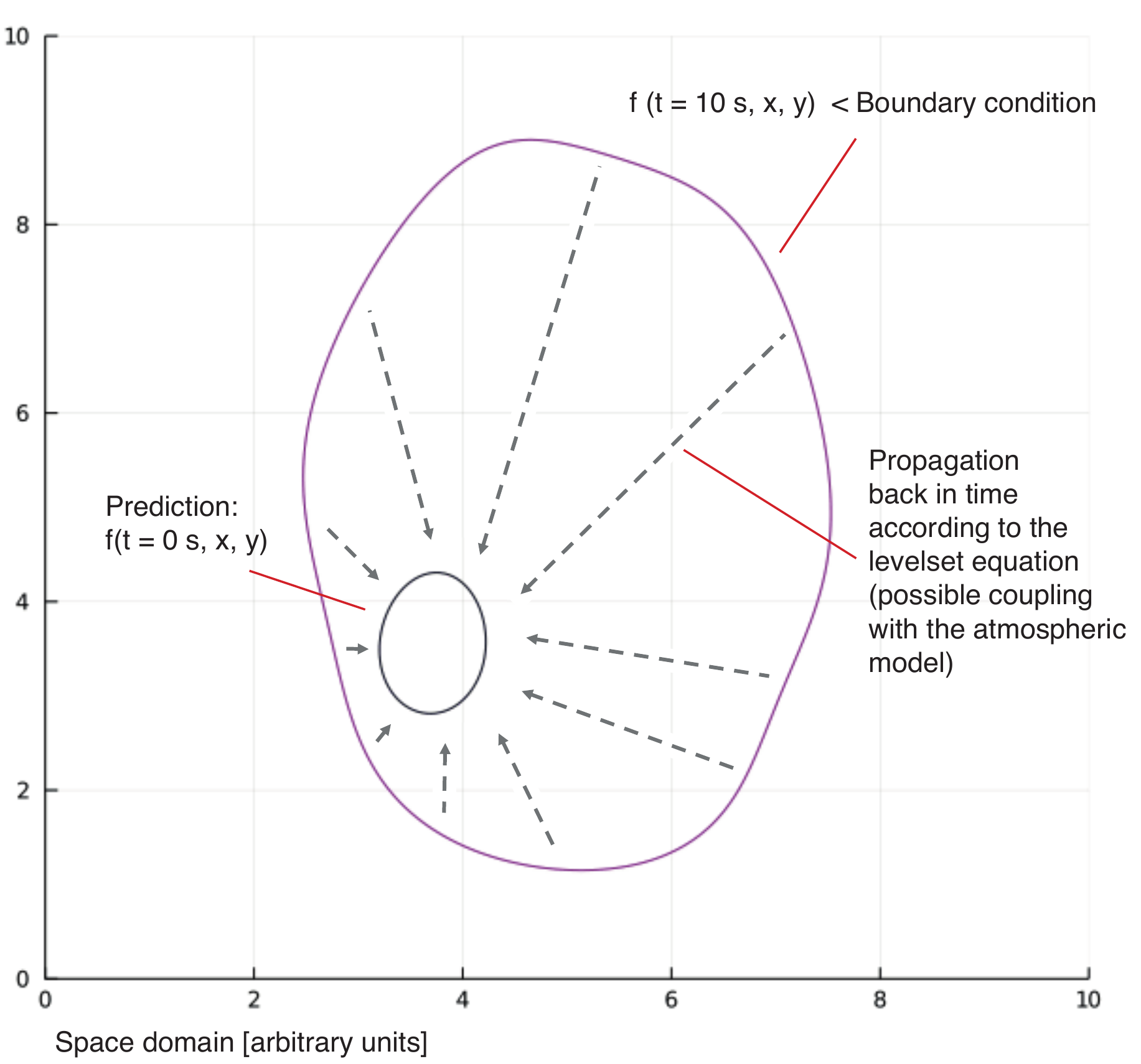}
\caption{Example of backward time prediction}
\label{fig:forensic}
\end {center}
\end{figure}

\vspace{-1pt}
A consequence of how boundary conditions are handled in our model is that, provided the fire shape in a certain instant, it would be possible to extrapolate the fireline in both "directions" of time. There is usually very little information about the temporal evolution of the fire (especially in remote areas), so the ability to extrapolate the shape of the spread backwards in time is crucial to understand where and how the fire started. These "forensic" investigations can provide insights that are valuable for fire prevention and protection. Our model is able to do so without any addition. The only requirement is that a "reasonable" level set is defined: as for the initial value problem, the level set equation must be set such that it does not have too big or too small values. We suggest to keep it in the range $(-20,20)$, however this is not necessary and is situation-dependent. It is very frequent that the negative part of the level set equation is indeed in the range $(-1,0)$. An example is portrayed in Fig.\ref{fig:forensic}.

\vspace{-10pt}

\section{\textbf{CONCLUSIONS}}
The study we carried out had the goal to investigate the applicability of the recently developed field of Scientific Machine Learning on climate, wildfire in particular, models. We have outlined some results that tell us that many improvements are needed in order to transform this into a validated product, but also show the big potential of our approach. We need to add further refinements to the implementation in order to carry out a precise time comparison between the two approaches, but the results obtained thus far show promising evidence.
The encouraging outcome inspires us to continue our work by improving the architectures and possibly employ them in different fields of research.
We hope that this line of research will be considered as a starting point for a more effective cohesiveness between Machine Learning and Physical Models in Climate Science, and thus further explored by other researchers.

\section{\textbf{ACKNOWLEDGEMENTS}}
This work was presented at the ProjectX 2020 competition by UofT AI.
We acknowledge University of Turin, Machine Learning Journal Club for supporting us.
We thank Professor Enrico Ferrero (Università del Piemonte Orientale), Professor Massimiliano Manfrin (University of Turin) and the whole Atmospheric Physics and Metereology Group, PhD Christopher Rackauckas (Massachussets Institute of Technology), PhD Kirill Zubov (Saint-Petesburg State University), Vaibhav Dixit (Julia Computing), PhD Brian Wee (Founder at Massive Connections), Dr. Rustem Arif Albayrak (NASA), PhD David Marvin (CEO at Salo Sciences), Professor Piero Fariselli (University of Turin) and Pietro Monticone M.Sc. student (University of Turin), for their precious help and availability.
We acknowledge the company \textit{Mollificio Astigiano} (Belveglio, Asti, Italy) for providing the computational power needed for this research and the\textit{ HPC4AI} center of the University of Turin for their support.

\clearpage  

\section{\textbf{APPENDIX}}\label{appendix}
\vspace{-5pt}

\begin{figure}[ht!]%[!ht]
\begin {center}
\includegraphics[width=0.4\textwidth]{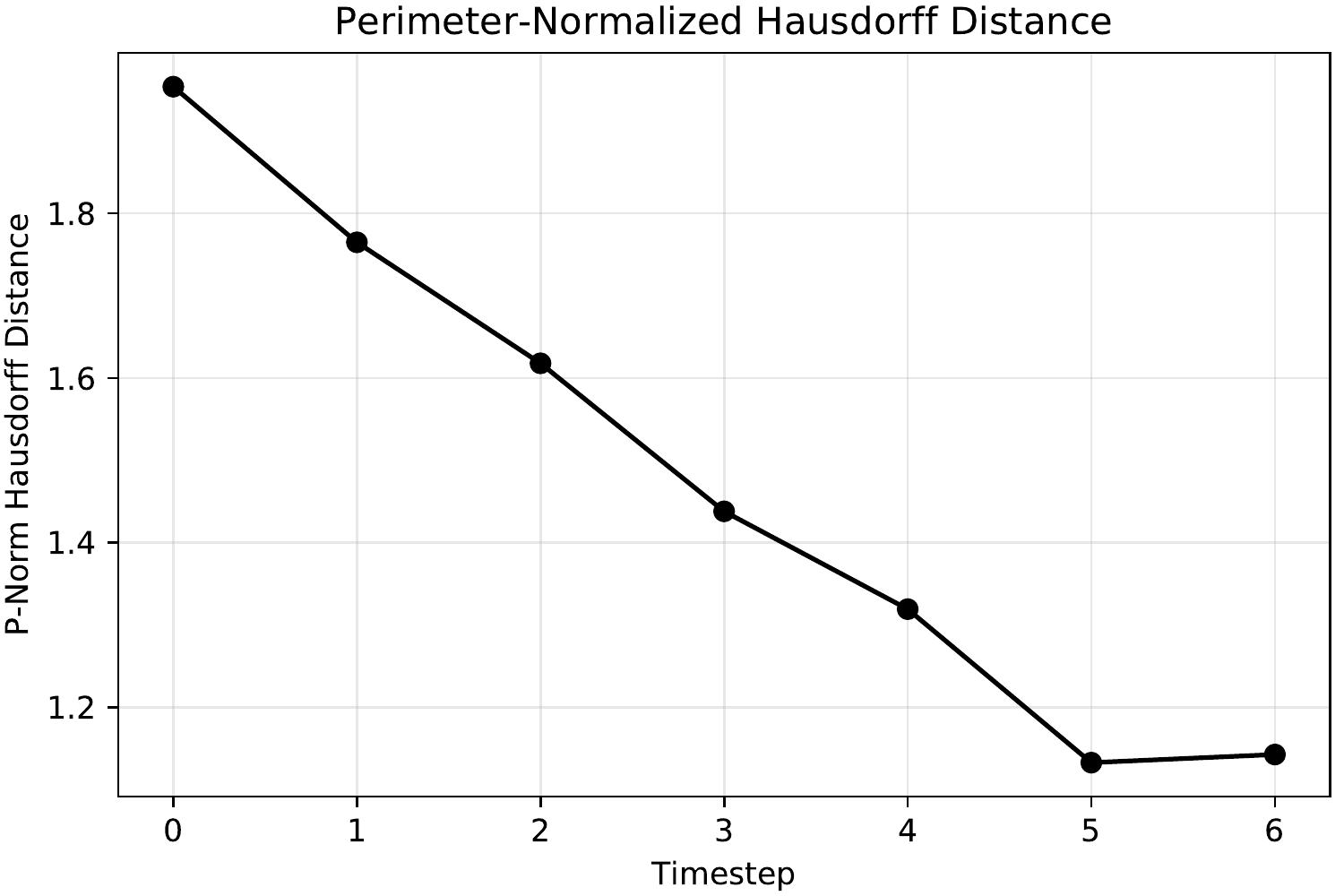}
\caption{Perimeter-normalized Hausdorff distance as a measure of the error between outputs - One Fire}
\label{fig:norm_perim_isom_h}
\end {center}
\end{figure}
\vspace{18pt}

\begin{figure}[ht!]%[!ht]
\begin {center}
\includegraphics[width=0.4\textwidth]{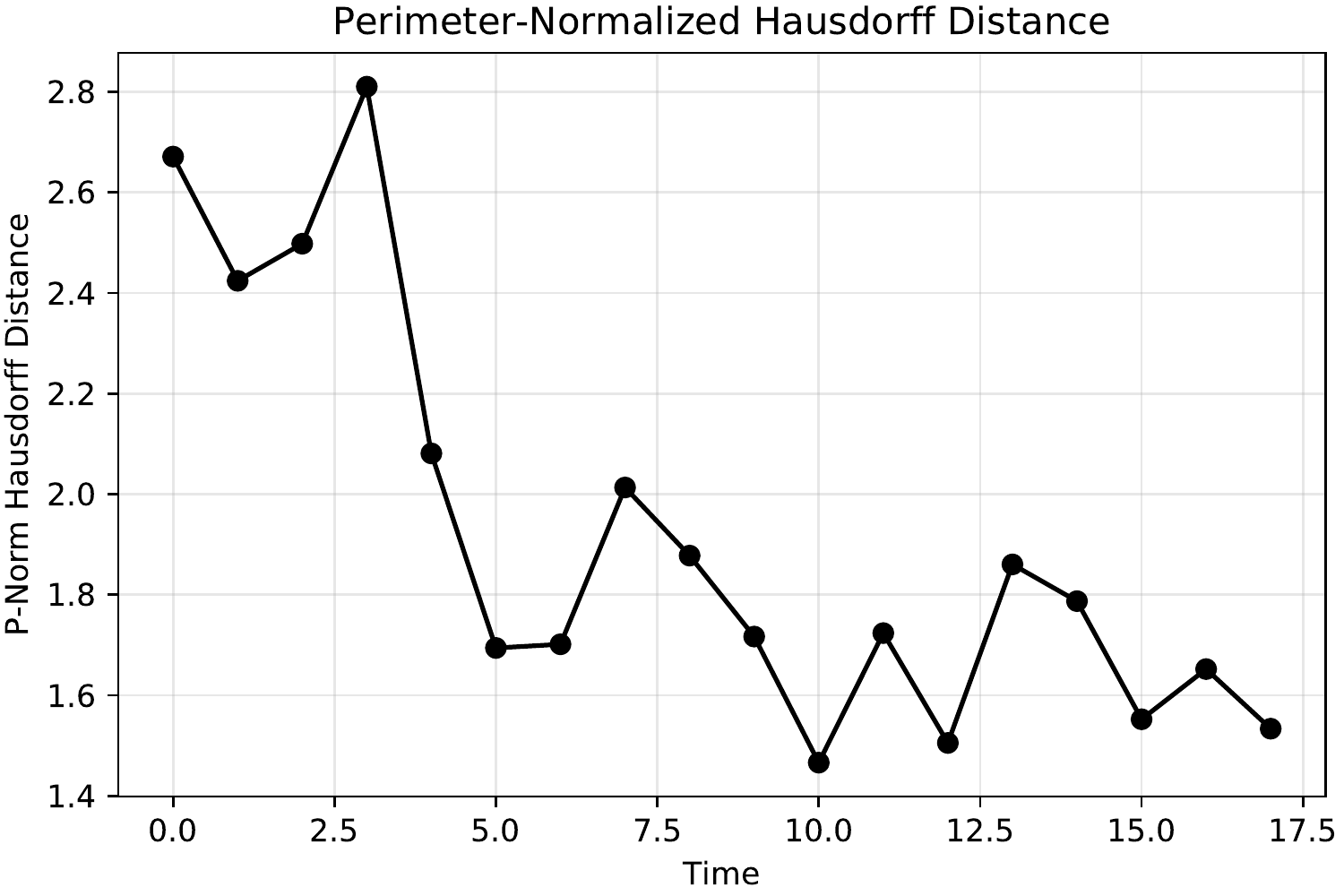}
\caption{Perimeter-normalized Hausdorff distance as a measure of the error between outputs - Isom Creek}
\label{fig:norm_perim_isom_h}
\end {center}
\end{figure}
\vspace{21pt}

\begin{figure}[ht!]%[!ht]
\begin {center}
\includegraphics[width=0.4\textwidth]{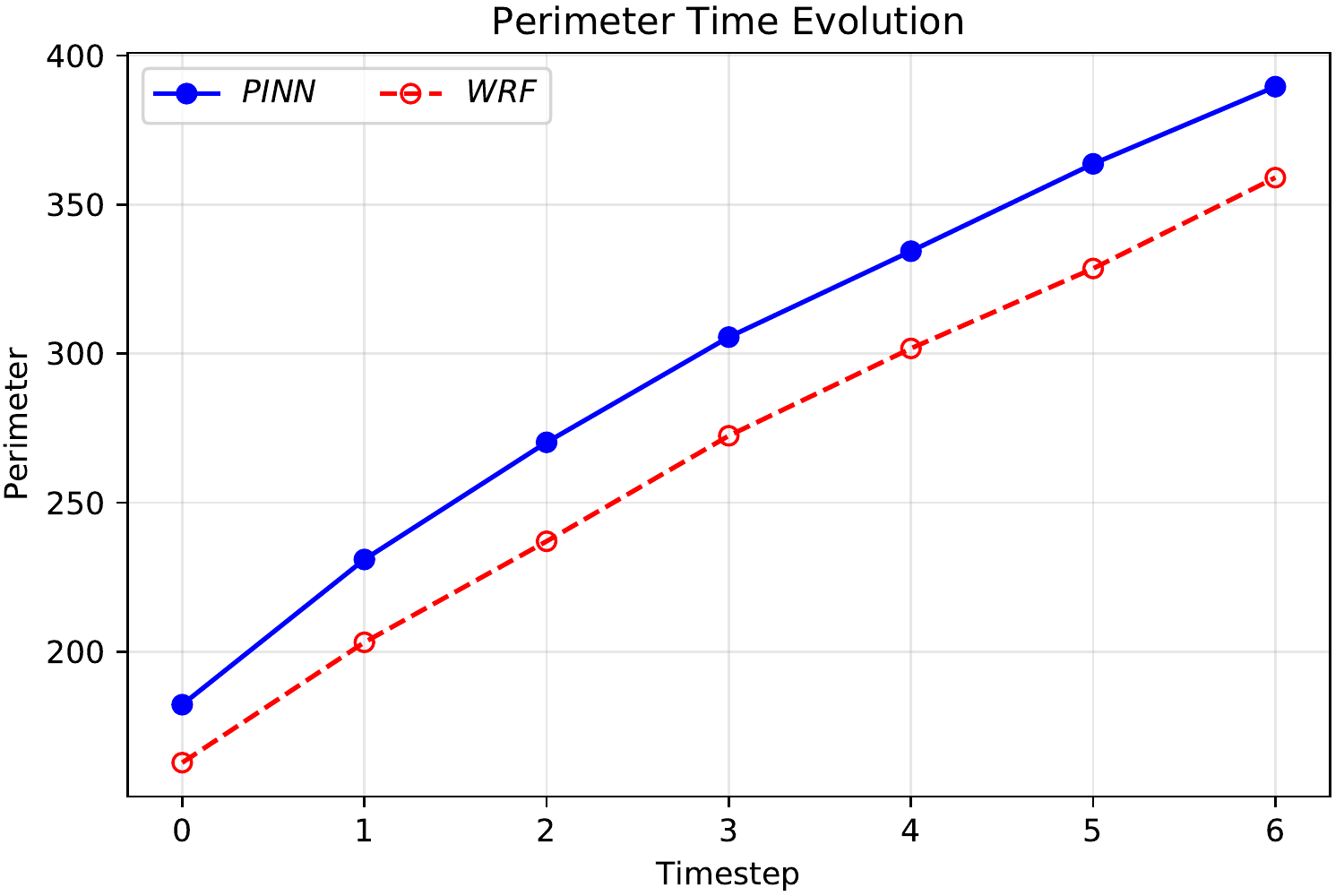}
\caption{Time evolution of the perimeters of the firelines - One Fire}
\label{fig:perimeters_one_fire_h}
\end {center}
\end{figure} 

\begin{figure}[ht!]%[!ht]
\begin {center}
\includegraphics[width=0.4\textwidth]{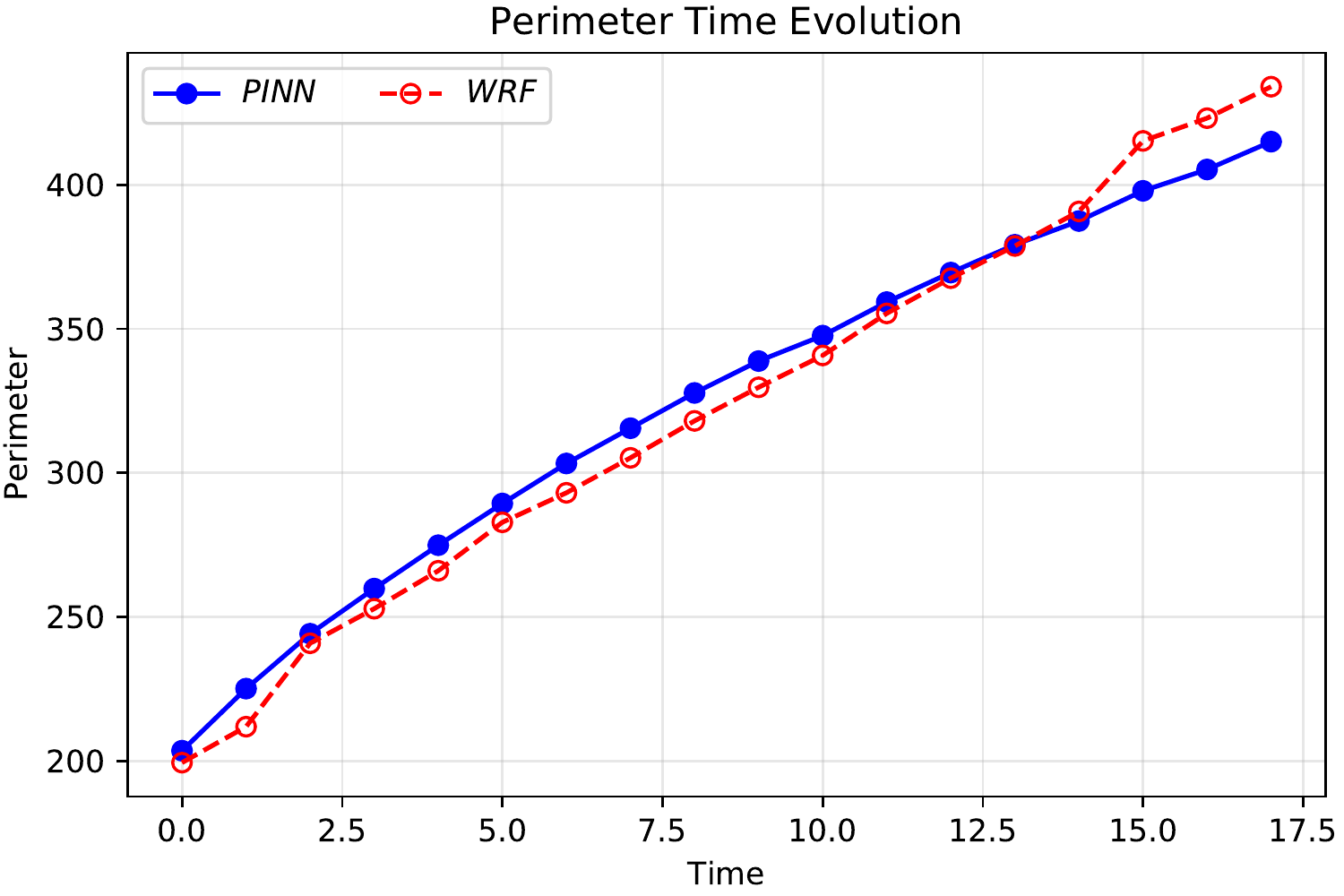}
\caption{Time evolution of the perimeters of the firelines - Isom Creek}
\label{fig:perimeters_isom_h}
\end {center}
\end{figure}

\begin{figure}[ht!]%[!ht]
\begin {center}
\includegraphics[width=0.4\textwidth]{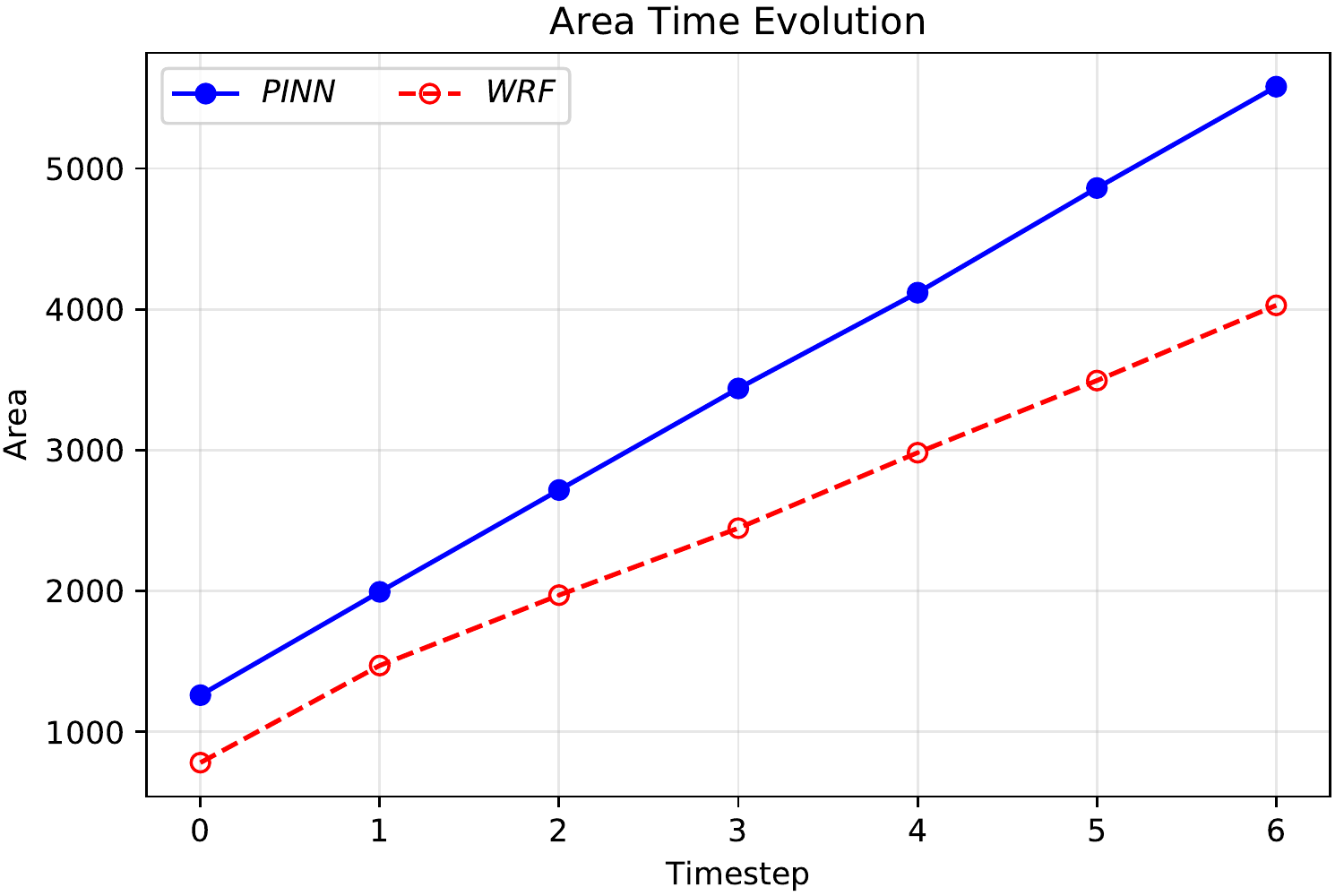}
\caption{Time evolution of the areas of the firelines - One Fire}
\label{fig:areas_isom_h}
\end {center}
\end{figure}

\begin{figure}[ht!]%[!ht]
\begin {center}
\includegraphics[width=0.4\textwidth]{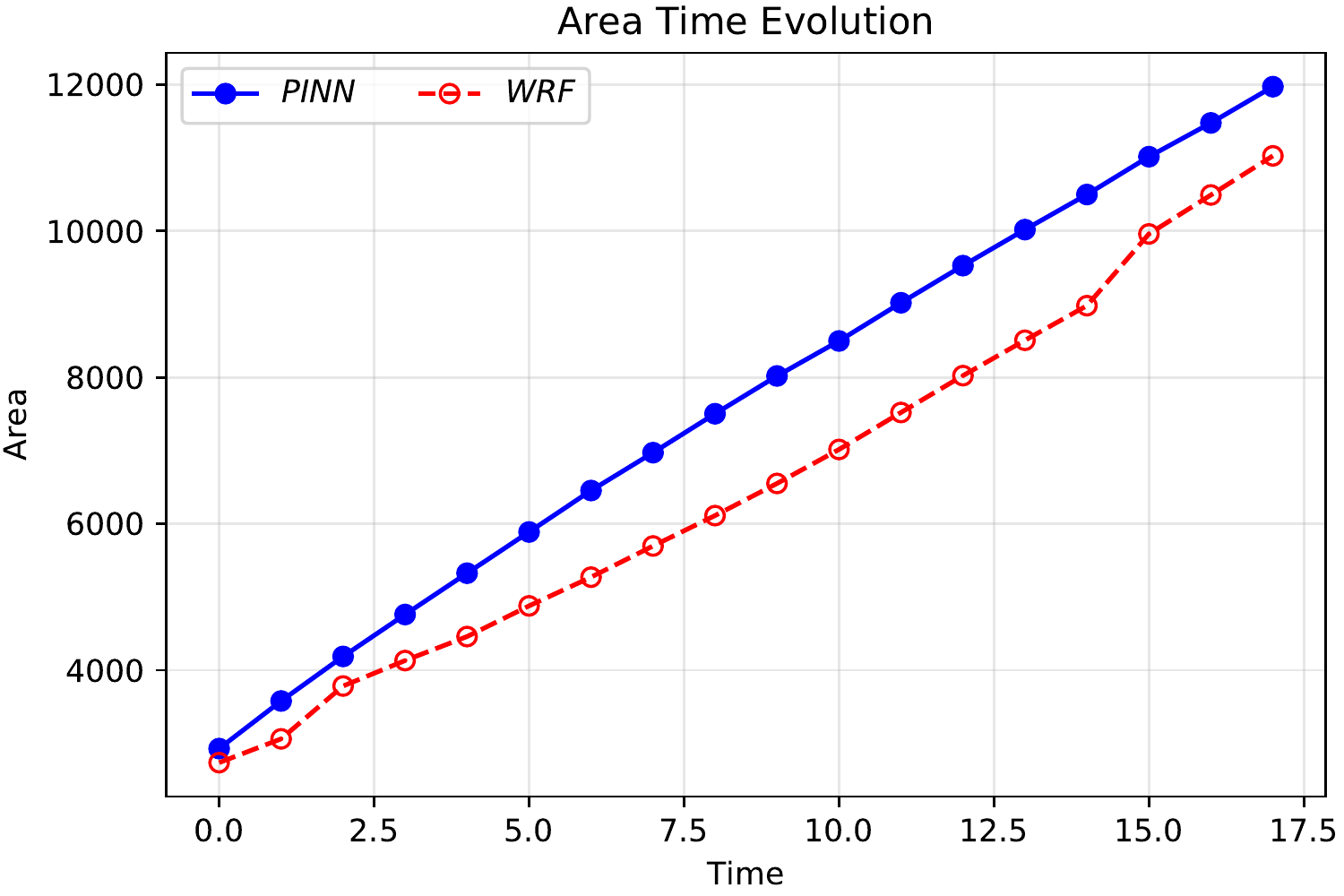}
\caption{Time evolution of the areas of the firelines - Isom Creek}
\label{fig:areas_isom_h}
\end {center}
\end{figure}

\begin{figure*}[t]%[t]
\centering
\includegraphics[width=1\textwidth]{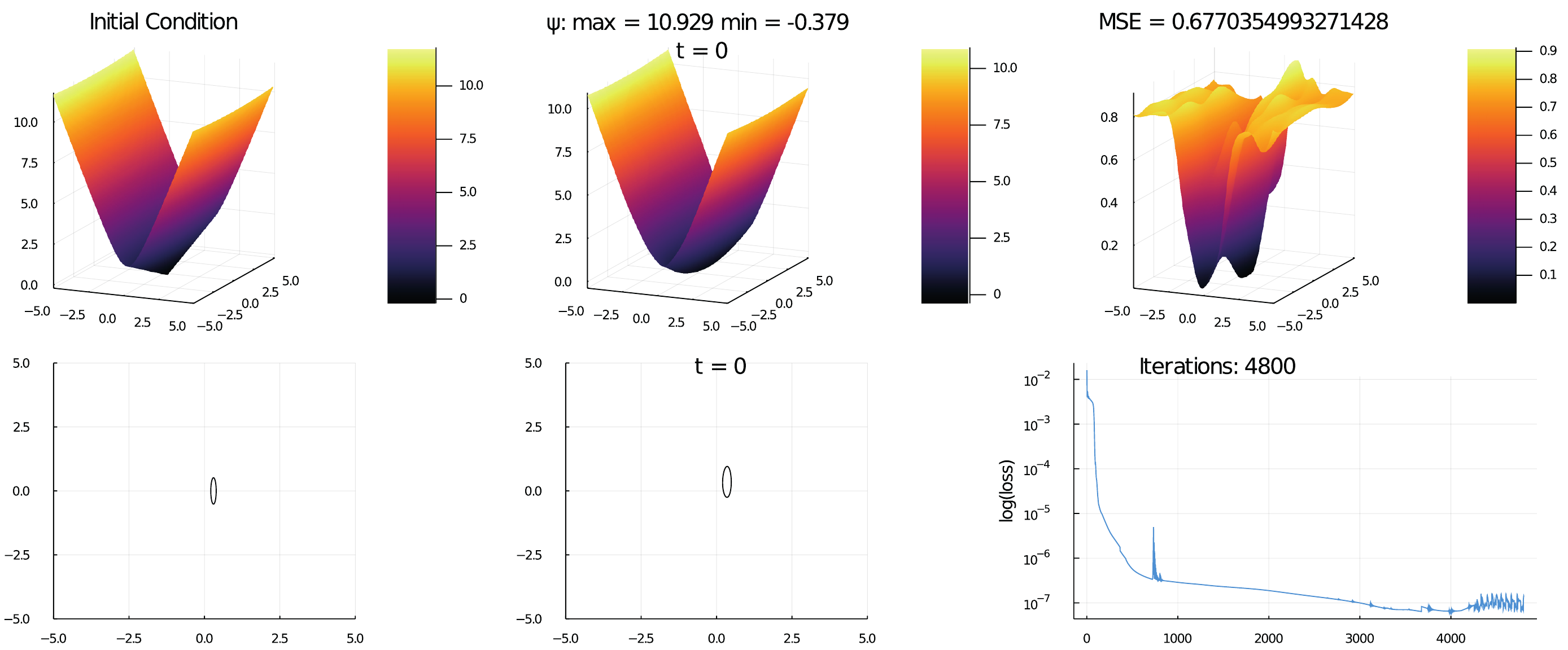}
\caption{Boundary condition comparison and training - One Fire}
\label{fig:boundary_one_fire}
\end{figure*}

\begin{figure*}[t]%[t]
\centering
\includegraphics[width=1\textwidth]{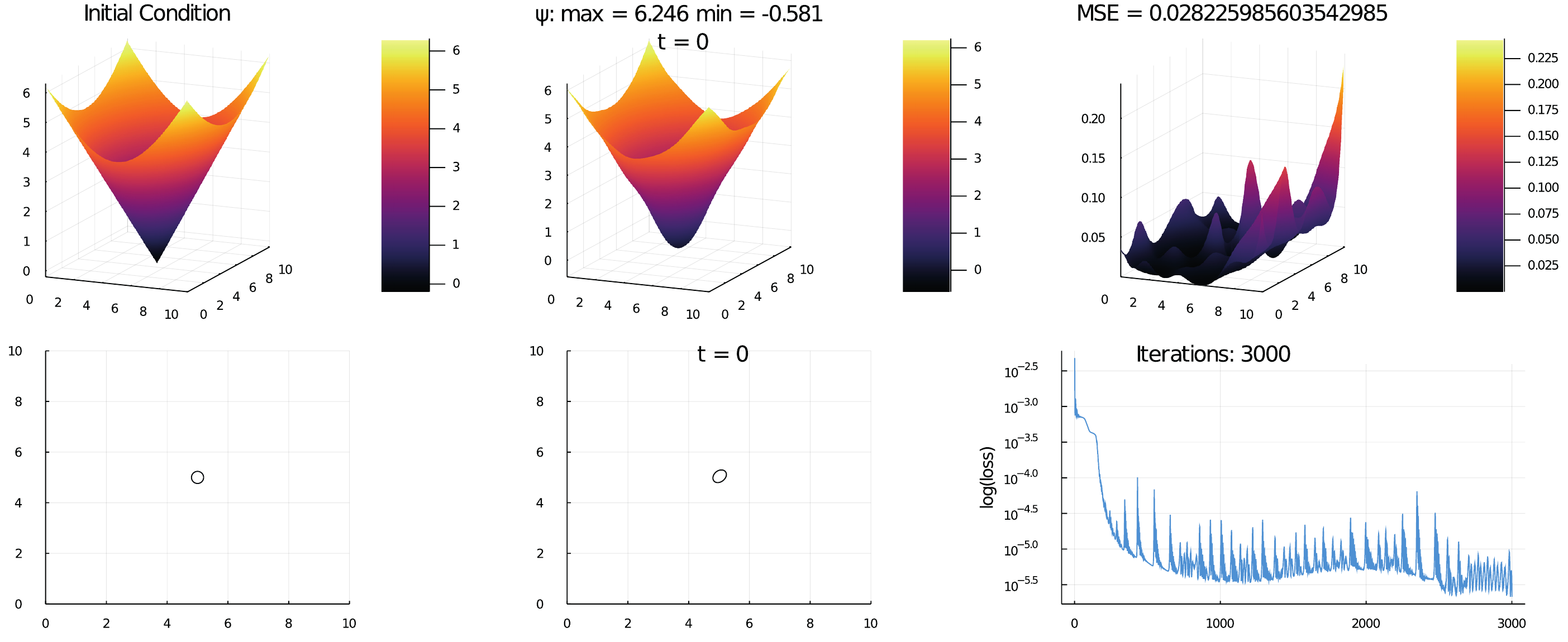}
\caption{Boundary condition comparison and training - Isom Creek}
\label{fig:boundary_isom}
\end{figure*}

\clearpage

\clearpage
\printbibliography
\end{document}